\documentclass[letterpaper]{article}
\usepackage[letterpaper]{geometry}
\usepackage{graphicx}
\usepackage{booktabs}
\usepackage{amsmath,amssymb}
\usepackage{mathtools}
\usepackage{adjustbox,placeins}
\usepackage[table*]{xcolor}
\usepackage{comment}
\usepackage{enumitem}
\usepackage{array}
\usepackage{algorithm}
\usepackage{algorithmicx}
\usepackage{newfloat}
\usepackage{hyperref}
\usepackage{listings}
\usepackage{titlesec}
\usepackage{etoolbox}
\usepackage{times}
\usepackage{epsfig}
\usepackage[noend]{algpseudocode}
\usepackage{colortbl}
\definecolor{Gray}{cmyk}{0.1,0.1,0.1,0.1}
\definecolor{lightgreen}{RGB}{144,238,144}
\newcolumntype{H}{>{\setbox0=\hbox\bgroup}c<{\egroup}@{}}

\newcommand{\ie}{{\em{i.e.}}}

\title{\textbf{PrivacyGAN: robust generative image privacy}}
\author{
    \textbf{Mariia Zameshina}\\
    \small{Univ Gustave Eiffel, CNRS, LIGM} \\ \small{F-77454 Marne-la-Vallee, France}\\
    \small{mariia.zameshina@esiee.fr}
    \and
    \textbf{Marlene Careil}\\
    \small{LTCI, Telecom Paris,} 
    \\\small{Institut Polytechnique de Paris}
    \and
    \textbf{Olivier Teytaud}\\
    \small{TAO, CNRS - INRIA - LRI}\\
    \and 
    \textbf{Laurent Najman}\\
    \small{Univ Gustave Eiffel, CNRS, LIGM} \\ 
    \small{F-77454 Marne-la-Vallee, France}
}

\date{}

\begin{document}

\maketitle

\begin{abstract}
Classical techniques for protecting facial image privacy typically fall into two categories: data-poisoning methods, exemplified by Fawkes, which introduce subtle perturbations to images, or anonymization methods that generate images resembling the original only in several characteristics, such as gender, ethnicity, or facial expression.

In this study, we introduce a novel approach, PrivacyGAN, that uses the power of image generation techniques, such as VQGAN and StyleGAN, to safeguard privacy while maintaining image usability, particularly for social media applications. Drawing inspiration from Fawkes, our method entails shifting the original image within the embedding space towards a decoy image.

We evaluate our approach using privacy metrics on traditional and novel facial image datasets. Additionally, we propose new criteria for evaluating the robustness of privacy-protection methods against unknown image recognition techniques, and we demonstrate that our approach is effective even in unknown embedding transfer scenarios. We also provide a human evaluation that further proves that the modified image preserves its utility as it remains recognisable as an image of the same person by friends and family.

\end{abstract}

\maketitle

\section{Introduction}

\begin{figure*}[htbp] 
\begin{center}
\includegraphics[width=0.7\textwidth]{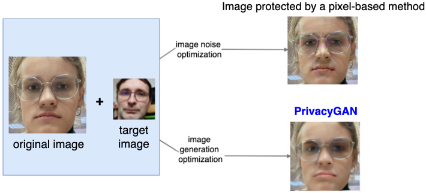}
\end{center}
\caption{\label{privacy_methods} Schema for both data poisoning and generative privacy methods.
We take the original image (OI) and create an image that is recognisable by human observers while being unlikely to be identified by image recognition methods using\\
(i) the classic approach by adding pixel noise such that it makes the modified image in the embedding space closer to the target image than to the original image,\\
(ii) (our approach, PrivacyGAN): generation of visually similar but distant images in the embedding space.}
\end{figure*}

Individuals often share personal photos on various social media platforms, which facilitates communication and connection with family, friends, colleagues, and customers. Unfortunately, a significant drawback of this practice is that it can sometimes be possible to identify individuals social media accounts by taking their picture in public \ref{sample2019facial} or by comparing their dating app photos to their business-related social media profiles. This is often made possible by the existence of datasets collected by scraping social media platforms. While face detection systems can be used by the government for criminal identification purposes, they also present opportunities for both internal and external misuse \ref{weise2009face}, including enabling stalkers to track their victims \ref{harwell2022facial}. Consequently, sharing real facial images publicly over the internet may compromise users privacy.

Multiple initiatives are dedicated to enhancing image and video privacy on the internet. One of the prominent groups is centred around \textbf{ anonymization methods}, which involve altering users pictures to resemble those of other individuals. For instance, Kim et al. \ref{Kim:2019} proposed a privacy-preserving adversarial protector network (PPAPNet) as an image anonymization tool. PPAPNet transforms an image into another synthetic yet realistic image while remaining immune to model inversion attacks \ref{wang2021variational}. Anonymization techniques may also preserve key characteristics such as background, emotions, and facial feature movements \ref{hellmann2023ganonymization} \ref{hukkelaas2019deepprivacy} \ref{barattin2023attribute} \ref{hukkelas23DP2}. These techniques are valuable when the objective is to maintain realistic appearances without the need to recognise individuals in photos or videos. Such approaches are particularly useful, for example, for maintaining anonymity while expressing opinions on video-sharing platforms. For instance, \ref{Gafni:2019} achieves the objective of decorrelating the identity while retaining the perception (pose, illumination, and expression). Some of these methods change only parts of the face. As an example, in their work \ref{qiu2022novel}, the authors suggest utilising generative techniques to enhance images that have been intentionally blurred or have had the subject's eyes obscured beforehand.

In the overview conducted by Wenger et al. \ref{wenger2023sok}, the authors tackle a challenge in the design of Anti-Facial Recognition (AFR) systems: finding a balance between \textbf{ privacy, utility, and usability }. They categorise AFR systems based on their target components, ranging from data collection and model training to run-time inference, all with the shared objective of thwarting successful recognition by unauthorised or unwanted models. Moreover, the authors stress the user preference for privacy tools with minimal overhead, a concept underscored by studies such as Sharif et al. \ref{sharif2016accessorize} and Dabouei et al. \ref{dabouei2019fast}. These findings highlight the significance of delivering protection against image recognition systems while mitigating any adverse effects on the user experience, a goal that many present anonymization methods struggle to attain. While certain attributes of images, such as gender, ethnicity, and facial expressions, can be retained through specific anonymization techniques, the resulting modified images frequently lack practicality for users. As a result, even though these images maintain crucial visual characteristics, the individual's identity within them may undergo substantial alterations, ultimately rendering them unidentifiable to acquaintances and family members.

In an effort to achieve a balance between utility and privacy protection, another area of research focuses on \textbf{ altering or obscuring facial images to maintain human recognizability while creating difficulties for neural networks to decipher}. Generally, these methods involve introducing precisely crafted pixel noise, causing the neural network to misclassify the image. These pixel-level perturbations have effectively challenged diverse image recognition neural networks. The dilemma of balancing privacy maintenance with recognition assurance of data-poisoning methods like Fawkes \ref{shan2020Fawkes} and Lowkey \ref{cherepanova2021Lowkey} is discussed in detail in \ref{radiya2021data}.

As an alternative approach to safeguarding images against unauthorised identification while preserving their utility for the users, one can consider \textbf{adversarial examples}. \ref{qiu2022novel} demonstrated that makeup transfer can be an effective means of countering various face recognition systems. However, this method has limitations, as the model's performance may be inconsistent between male and female images due to an imbalance in the makeup transfer training dataset. This method is also ineffective in cases where the face cannot be found on an image, as it is not possible to transfer makeup in this case. Additionally, some individuals may find the use of makeup transfer images unacceptable.

\textbf{In the present paper, we propose a general approach to the use of generative methods for privacy.} We train our methods to be effective for embedding methods, on which pixel-based methods such as Fawkes used to fail. Our goal is to create images that resemble original photographs and are suitable for sharing on social media platforms, while also preventing identification by modern image recognition neural networks without using anonymization. To achieve this, we explore the effectiveness of two generative methods: a generative adversarial neural network StyleGAN \ref{karras2019style}, and the autoencoder VQGAN \ref{esser2021taming}. These generative methods are known for their realistic image generation capabilities, which adds an added layer of difficulty for neural network recognition.

By modifying facial images using generative methods, we aim to preserve their recognizability to human observers while rendering them unrecognisable to many existing image recognition neural networks. Inspired by pixel-based methods like Fawkes \ref{shan2020Fawkes}, we propose modifying the generated ``private'' images towards a different target image in the embedding space and evaluate the robustness of our approach against unknown image recognition neural networks. We validate our privacy methods on the Labeled Faces in the Wild (LFW) dataset \ref{huang2008labeled}, as well as introduce a new dataset of face crops extracted from Casual Conversations \ref{hazirbas2021towards} to ensure their effectiveness in various environments. In Fig. \ref{privacy_methods} we present the schema of image modification using both pixel-based and generative methods.

Our proposed generative tools make subtle modifications to user images without adding pixel noise, so the resulting photos look natural and protect user privacy. Our algorithms operate within a black-box framework and demonstrate their efficacy against image recognition techniques they were not specifically trained on. We offer flexibility in selecting methods and privacy settings and conduct a comparison between our approach and existing state-of-the-art privacy protection methods, such as Fawkes.

{\textbf{To sum up the claims of the present paper}}, we

\begin{enumerate}

\item propose a novel approach to facial image privacy based on generative methods;

\item create a new privacy evaluation approach based on the percentage of dataset images that are closer in an embedding space to a modified ``private'' image than to an original image;

\item propose a new facial image dataset extracted from the Casual Conversations dataset \ref{hazirbas2021towards} videos;

\item evaluate the privacy of the modified images against various embedding methods {(including transfer to embeddings not used in our privacy method)} and provide human evaluation of image quality for state-of-the-art and novel privacy methods.

\end{enumerate}

\section{Privacy Algorithms}
\subsection{A well-known pixel-based method: Fawkes.}
Fawkes \ref{shan2020Fawkes} is a data poisoning method that presents subtle image perturbations to the images. One of its main features is the concept of target image: by suggesting elements from a side target image, Fawkes {ensures} that the modified image will be recognised as another person by neural networks, thus ensuring privacy. Unlike just maximising the distance between the embedding of the original image and the modified image, this method

\begin{enumerate}
\item helps to keep the embedding of modified images within a valid range for a given dataset and
\item ensures that the embedding of the modified image does not stay close to its original version in the given dataset.
\end{enumerate}

The idea of Fawkes is to pair each original image (\textbf{OI}) with a target image (\textbf{TI}). Then Fawkes associates to each original image a ‘cloak image’ (\textbf{CI}) which consists of noise obtained by optimising the following loss:
$$L_{Fawkes}= ||emb(TI) - emb(OI \oplus CI)||,$$
where: \textbf{OI} is an original image; $\oplus$ is capped addition; $emb$ is an embedding method used for cloak optimisation in order to obtain a modified ``private'' version of an original image; \textbf{TI} is a target image; the private version of OI should be labelled the same as TI by the chosen image recognition system;
$\rho$ is a parameter that caps the noise strength;
\textbf{CI} $ < \rho$ is a cloak or a noise that should be added to OI in order for it ensure its' privacy;
$OI \oplus CI$ is the published rendition of $OI$. In our experiments, by default, we use the ``high'' mode of Fawkes (as mentioned in \ref{fawkesgithub} and \ref{shan2020Fawkes}), since it provides a decent level of protection, and it is possible to compare this setting of Fawkes with our methods.

\subsection{Our proposed generative methods based on VQGAN and StyleGAN}
Our proposal involves utilising generative models for privacy protection, with a focus on generating an image that closely resembles the original in visual appearance while safeguarding users against image recognition attacks. Our objective is not to anonymize the image. Another generative method that transfers makeup in order to protect facial privacy (AMT-GAN) is reviewed in the supplementary material.
In this paper, we expand on the idea of target images introduced in Fawkes. We use target images to ensure that the modified version of an image is closer to the target image than the original image in the chosen embedding space.
To select target images, we choose from images in the dataset that have not been used in experiments. For each specific image, we select a target image based on its distance from the original image in the chosen embedding method used for optimization. The chosen target image should be far enough from the original image to ensure effective privacy protection.

We select the loss function $L$ based on the goal of preserving the identity of the original image (OI) for humans while ensuring that the generated image (GI) embedding is as close as possible to the distant target image embedding. For this purpose, we use the Learned Perceptual Image Patch Similarity (LPIPS) distance for the preservation of OI identity and optimise the embedding distance between the generated image (GI) and the target image (TI) to achieve the closest possible embedding.

The loss for any generated image always consists of the sum of the following parts:

\begin{enumerate}
\item LPIPS distance between generated and original image
\item For each of the embeddings used for optimisation, coefficient $K$ multiplied by the mean squared distance between the modified ``private'' image embedding and target image embedding.

$$L_{\overset{generative}{privacy}} = LPIPS(OI, GI) + $$
$$ + K \times \sum_{\mathclap{\substack{emb \in\, embeddings}}} \vert\vert \,emb(GI) - emb(TI) \vert\vert.$$

\end{enumerate}

The hyperparameters of {the loss described} are the coefficient $K$ (\ie, weight compared to LPIPS) for the embedding distance, the learning rate, the batch size, and the number of iterations.

\subsection{Generative Privacy Algorithm: PrivacyGAN}
Here, we describe PrivacyGAN, an algorithm that we propose for creating private versions of facial images.

\begin{algorithm}[h!]
\begin{algorithmic}
\Require $OI$: Original Image
\Ensure $GI$: Generated Image
\State $TI \gets$ Target Image \Comment{(distant from OI in the embedding space)}
\State $z \leftarrow$ random
\State $G \leftarrow$ image generation method
\State $K \leftarrow$ optimisation coefficient
\State $chosen\_embeddings \leftarrow$ list of embedding methods for optimisation \Comment{(we distance GI from OI in these embedding spaces)}

\For{i in range(0, num\_iterations)} 
\State $GI \gets G(z)$ 
\State $emb\_dist\leftarrow 0$
	\For{$emb$ in {$chosen\_embeddings$}}
	    \State $emb\_dist \, \mathrel{\raisebox{0.19ex}{$\scriptstyle+$}\!\!=} ||emb(GI)-emb(TI)|| $
	\EndFor
	   \!\!\! \!\!\!\!\!\!\!\!\State $lpips\_dist \!\gets \!\!LPIPS(GI, OI)$
	    \State $loss \gets lpips\_dist + K \cdot emb\_dist$
 	    \State $z \gets update(z,loss, \nabla_{\mathrm{loss}})$
\EndFor
\Return  $GI$ 
\end{algorithmic}
\caption{\label{alggen}PrivacyGAN: Private image generation algorithm}
\end{algorithm}

As input to the algorithm, we use an original image \textbf{OI} and a target image \textbf{TI}. The algorithm aims to produce an image \textbf{GI} which would be a ``private'' {version of OI, unrecognisable by many image recognition neural networks.}
In order to do that, we are using generative {methods} such as StyleGAN and VQGAN.
TI is chosen randomly among the images furthest in embedding space from OI.

The algorithm consists of an iterative optimisation process, where $\textbf{num\_iterations}$ {represents the} number of iterations and $\textbf{G}$ is {the} image generation method. In the latent space of $G$, we find a latent variable $\textbf{z}$ and generate {an} image G(z), {which we refer to as GI}. For each of the embedding methods in the set \textbf{chosen\_embeddings}, in each iteration of the algorithm, we compute the embedding distance between GI and TI, as well as the LPIPS distance \ref{lpips} between images GI and OI. We use the computed distances {to calculate the} $L_{\overset{generative}{privacy}}$ that we mention as 'loss' in the algorithm.

\section{Evaluation Methods}
\subsection{Metrics for Privacy}
In order to evaluate the privacy of generated images, it is important to determine how far the generated image is from the original in the dataset. After applying an image recognition neural network, attackers may choose to verify if the person on the image matches the top few possible results. That is why the method would work better for privacy protection if: i) the modified image would not be recognised as its original version; and
ii) the original image would be far away from the modified one in the embedding space.

We ensure ii) not only by using existing evaluation methods such as {Recall@}$k$ that help us to make sure that the modified ``private'' image is far from the original in absolute values but also by introducing a novel evaluation method that verifies the original and modified image being far in embedding space relative to the dataset size.

To measure the distance from the original image to its modified private version, we use the following privacy metrics: {Recall@}$k$ and $Percentage$.

{Recall@}$k$ for the set of query images $L$ (which can either be original or modified images) and test images $M$, is defined as 
$$ Recall(L,M,k) = 100 \frac{\sum\limits_{q \in L} 1_{Id(q) \in Id(N(q, k, M))}}{||L||},$$ 
where
function $Id$ maps the set of people’s images to the set of (unique) identities of the individuals present on these images,
function $N(q, k, M)$ returns a set of $k$ images from $M$ that have the closest embedding to the one of the query image $q$.

We propose the use of a new metric, called the ``Percentage'', in addition to the Recall metric, to evaluate the effectiveness of our privacy methods. The reason for introducing this new metric is to ensure that the modified image is not only far from the original image in absolute terms, as ensured by {{Recall@}$k$}, but also in terms of the percentage of dataset size. This provides a common privacy metric that can be used to compare the effectiveness of our methods across different dataset sizes.

$Percentage$ is the proportion of images for each query image from the dataset $L$ in between the query image and the closest image with the same identity from the dataset $M$: 
$$Percentage(L,M,k) = 100 \sum\limits_{q \in L} \frac{Between(q, N(q, 1, M))}{||L|| \times ||M||},$$
where the function $Between(q_1, q_2)$ returns the number of images in the dataset $M$ that have a smaller distance to the embedding of $q_1$ than the distance in-between the embeddings of $q_1$ and $q_2$.

\subsection{The problem of transfer}

In practical scenarios, it is crucial that privacy methods are effective against various image recognition neural networks. We optimise our privacy methods to be effective for specific embeddings, and transferring to a different embedding method can be challenging as new methods are continually emerging. It is impossible to guarantee that privacy methods will be effective against future attacks, as some methods have been broken by newer recognition neural networks \ref{radiya2021data}. To evaluate the effectiveness of our proposed methods, we conducted two sets of optimisation experiments.

\begin{table*}[t]
	\begin{center}
	    \begin{adjustbox}{max width=0.87\textwidth}
		
		\begin{tabular}{@{}l|rrrr|rHHHH@{}}
               \rowcolor{gray!30}
                \multicolumn{1}{c}{} & \multicolumn{4}{|c|}{PrivacyGAN} & \multicolumn{1}{c}{Pixel-based} & \multicolumn{4}{c}{}\\
			  & StyleGAN & StyleGAN & VQGAN & VQGAN & Fawkes & F+S & F+V & S+F & V+F \\
			\rowcolor{gray!30} &&\_0.003\_500&&\_0.005\_128&&&&&\\
			\toprule
			Percentage  & 8.110 & 0.654 & \textbf{14.696} & 0.861 & 0.782 & 11.491 & 4.048 & \textbf{22.370} & 20.105 \\
			\rowcolor{gray!30 }Recall@1:m.i. & 1.754 & 22.085 & \textbf{0.047} & 19.242 & 20.521 & 1.280 & 5.403 & 0.806 & 1.043 \\
Recall@1:o.i. & 2.180 & 22.133 & \textbf{0.332} & 22.180 & 23.886 & 1.422 & 6.825 & 0.616 & 0.900 \\
			\rowcolor{gray!30}Recall@3: m.i. & 6.114 & 61.564 & \textbf{0.758} & 54.597 & 56.398 & 3.697 & 14.929 & 1.754 & 2.322 \\
			Recall@3: o.i. & 5.308 & 60.142 & \textbf{0.900} & 53.981 & 58.246 & 3.555 & 15.355 & 1.564 & 1.754 \\
			\rowcolor{gray!30}Recall@5: m.i. & 8.815 & 77.678 & \textbf{1.374} & 70.711 & 74.502 & 5.261 & 21.280 & 2.701 & 3.791 \\
			Recall@5: o.i. & 7.109 & 75.782 & \textbf{1.327} & 67.820 & 72.796 & 4.929 & 20.806 & 2.133 & 2.559 \\
			\rowcolor{gray!30}Recall@10: m.i. & 13.365 & 86.256 & \textbf{2.986} & 79.668 & 83.412 & 9.289 & 31.706 & 4.313 & 5.735 \\
			Recall@10: o.i. & 11.422 & 85.355 & \textbf{2.512} & 77.773 & 82.938 & 8.341 & 30.047 & 4.076 & 4.408 \\
			\rowcolor{gray!30}Recall@50: m.i. & 32.417 & 94.408 & \textbf{11.280} & 92.227 & 92.986 & 24.028 & 56.730 & 12.133 & 14.455 \\
			Recall@50: o.i. & 28.768 & 94.028 & \textbf{10} & 92.464 & 93.981 & 22.133 & 54.076 & 11.232 & 13.507 \\
			\rowcolor{gray!30}Recall@100: m.i. & 43.697 & 96.303 & \textbf{19.336} & 95.166 & 95.592 & 34.313 & 66.730 & \textbf{18.578} & 20.995 \\
			Recall@100: o.i. & 40.806 & 96.019 & \textbf{17.062} & 95.071 & 96.303 & 32.417 & 65.024 & \textbf{16.540} & 19.005 \\
			\bottomrule
		\end{tabular}
	    \end{adjustbox}

	\end{center}
\caption{\label{facenet_facenet} Test on the LFW dataset. Evaluation for the same embedding that was used for training (no transfer): PrivacyGAN (based on VQGAN or StyleGAN) is optimised with FaceNet, tested with FaceNet, and compared to Fawkes in ``high'' mode (meaning: high privacy). We see that PrivacyGAN equipped with standard versions of StyleGAN and VQGAN obtains better privacy results compared to Fawkes. \label{lfw-5-facenet-}
}
\end{table*}

The first experiment involves optimising StyleGAN and VQGAN image generation to be effective against the FaceNet embedding method. We aim to make the FaceNet embedding of the generated image distant from that of the original image during the optimisation process. We compare our proposed methods to Fawkes, which uses the same embedding method for optimisation, in Table \ref{lfw-5-facenet-}. Additionally, we aim to test the transferability of our methods to embedding methods introduced after FaceNet, which Fawkes does not prove to be effective against \ref{radiya2021data}.

The second experiment involves optimising StyleGAN and VQGAN image generation using MagFace and MobileFaceNet embedding methods, which have been shown to increase the robustness of generated images. We also compare them to the makeup transfer method AMT-GAN \ref{hu2022protecting}

\section{Datasets and embeddings}

\subsection{The Labelled Faces in the Wild}
The dataset \ref{huang2008labeled} contains multiple images for each person, with the number of images per person varying between $1$ and $530$. To ensure fairness, we extract a sub-dataset from the original dataset, which includes $5$ randomly chosen images per person. We exclude images of people who have less than $5$ photos present in the dataset. This sub-dataset is referred to as LFW in the following sections of this paper.

\subsection{The Casual Conversations dataset} The Casual Conversations dataset \ref{hazirbas2021towards} {comprises} $45186$ videos, each of which features one person. We select $997$ videos and extract $5$ face crops of size $456 \times 456$ per person present in the dataset (in case the video contains face crops of a required size).

The process of selecting these face crops is as follows:
\begin{enumerate}

\item We select all the time frames from the video featuring a specific person.
\item We check if, among these time frames, there are at least $5$ non-consecutive ($\pm 10$) time frames that satisfy the following conditions:
\begin{itemize}
\item they contain a face crop of a size at least $456 \times 456$ with a margin of size $100$;
\item average brightness of a time frame is at least $70$. This condition is required since, among the videos in the CC dataset, there are many that were recorded in complete darkness, and it is not realistic to have such face crops as profile pictures.
\end{itemize}
\item If there are more than $5$ time frames selected, we randomly choose $5$ of them and add them to the dataset.
\end{enumerate}

We make sure that we don't select successive frames of the video since they could contain identical face crops. 

For the confounders set, we randomly choose different people's face crops that also satisfy conditions 1 and 2 and do not feature a person who was already selected for our primary dataset before.

The key difference between our novel dataset and LFW is that faces in our proposed dataset have similar backgrounds and are taken within a short timeframe, creating an additional challenge for privacy protection. That lack of variety makes this particular dataset very interesting for our research. By testing our methods on it, we are able to ensure that, even if there are many very similar photos of the same person in the dataset, the proposed privacy tools can still be effective. It is particularly important in cases where people publish their images from similar locations on different platforms over the internet. Later, we refer to this dataset as \textbf{CC}.

The complete code for the face crop dataset extraction will be provided at the time of publication. 

\subsection{Our proposed methods for transfer to unknown embeddings: optimising on multiple embeddings \label{emb_methods}}

The embedding methods that we are using in this paper are the following: FaceNet \ref{schroff2015facenet}; ArcFace \ref{deng2019ArcFace}; SphereFace \ref{liu2017SphereFace}; MagFace \ref{meng2021MagFace}; MobileFaceNet \ref{chen2018mobilefacenets} with implementation from the FaceX-Zoo library \ref{wang2021facex}; and ResNet\_152 \ref{he2016deep} with implementation from the FaceX-Zoo library \ref{wang2021facex}.

We evaluate the effectiveness of our proposed privacy methods in a black-box setting (\ie, {robustness to unknown image recognition methods not used in the privacy method}). We optimise the generated image for one or two embeddings from the list and then check the generated image against all the other embeddings. Thus, we make sure that our privacy methods transfer well to unknown embeddings and can be used for the privacy protection of real photos published online.

\section{Experiments and Results}

Here we define the settings and notations that {we use} in our experiments.

We set the hyperparameters of the generative privacy loss ($L_{\overset{generative}{privacy}}$) for our experiments in the following way: the learning rate to $0.01$ and the batch size to $32$. The only parameters that we modify from experiment to experiment are the coefficient $K$ and the number of iterations.

We have introduced the notations ``o.i'' and ``m.i'' to represent the original image and the modified image generated by any privacy-preserving algorithm, respectively. For our recall evaluation, we select either the original image (o.i. context) or the modified image (m.i. context) as the query image, where the dataset used for recognition includes modified images and confounders for the former and includes the original images and confounders for the latter.

In all metric calculations, we also use a set of confounders, which are {not used as queries} in {our} experiments and are sourced from the same dataset as the original images. The number of confounders is always less than or equal to $\frac{1}{5}$ of the number of original images. To compare privacy protection methods, we use the average value of the transfer recall (\ie, Recall@10) for all embeddings for which the algorithm was not optimized. Incorporating confounders in our experiments brings us closer to real-world scenarios, where datasets may contain unrelated images that can potentially affect the experiment results.

Moreover, in order to compare VQGAN, StyleGAN, and Fawkes {to one another}, we use different sets of parameters chosen to match the transfer recall results. Thus, we are able to compare the generated image quality and see which of the methods generates the best images in terms {of image quality for a given privacy performance (measured by recall).}

Later in this section, we use the following notations: By \textbf{standard version of StyleGAN} we mean PrivacyGAN equipped with StyleGAN optimized with a coefficient $K=0.03$ for embedding distance in the loss and $128$ iterations;
By \textbf{standard version of VQGAN} we mean PrivacyGAN equipped with VQGAN optimized a coefficient $K=0.03$ for embedding distance in the loss and $1000$ iterations;
By StyleGAN\_$x$\_$y$ / VQGAN\_$x$\_$y$ we mean PrivacyGAN equipped with StyleGAN/VQGAN optimized a coefficient $K=x$ for embedding distance in the loss and $y$ iterations.

\subsection{ Experiment 1: Comparing Pixel-Based and Generative Methods Optimised for One Embedding on the LFW Dataset}
In Table \ref{lfw-5-facenet-}, we compare standard versions of VQGAN and StyleGAN and their versions StyleGAN\_0.003\_500 and VQGAN\_0.005\_128 that we prepare specifically to match the privacy results of Fawkes. With this parametrization, they have the same transfer recall score, which allows us to compare {fairly the} image quality of our proposed generative methods and the state-of-the-art pixel-based method Fawkes.
{The transfer recall values (average values of {Recall@10} for other methods from the list) are $62.16\%$ for StyleGAN, $89.28\%$ for StyleGAN\_0.003\_500, $65.49\%$ for VQGAN, $90.07\%$ for VQGAN\_0.005\_128, $90.90\%$ for Fawkes. }
In order to make sure that image privacy is robust against various facial recognition systems, we study a transfer to different embedding methods (ArcFace, MagFace, SphereFace, MobileFaceNet, ResNet\_152).
Some results are in Table \ref{st_vq_facenet_sphereface} (SphereFace) and in Table \ref{st_vq_facenet_magface} (MagFace).

More results can be found in the supplementary material.

\begin{table*}[t]
	\begin{center}
	    \begin{adjustbox}{max width=0.87\textwidth}

		\begin{tabular}{@{}l|rrrr|rHHHH@{}}
  \rowcolor{gray!30}
 \multicolumn{1}{c}{} & \multicolumn{4}{c}{PrivacyGAN} & \multicolumn{1}{c}{Pixel-based} & \multicolumn{4}{c}{}\\
			  & StyleGAN & StyleGAN & VQGAN & VQGAN & Fawkes & F + S & F + V & S + F & V + F \\
			 \rowcolor{gray!30}&&\_0.003\_500&&\_0.005\_128&&&&&\\
			\toprule
			Percentage  & \textbf{5.181} & 1.388 & 5.104 & 1.271 & 1.213 & 6.809 & 8.143 & \textbf{10.430} & 10.078 \\
 
			\rowcolor{gray!30} Recall@1: m.i. & 9.526 & 21.896 & \textbf{ 8.768} & 21.043 & 21.611 & 6.967 & 4.360 & \textbf{3.649} & 4.218 \\
			Recall@1: o.i. & 9.431 & 22.891 & \textbf{8.578} & 23.223 & 23.791 & 7.820 & 6.872 & \textbf{4.550} & 5.877 \\
 
			\rowcolor{gray!30} Recall@3: m.i. & 21.422 & 53.744 & \textbf{20.142} & 54.929 & 55.877 & 16.398 & 10.853 & \textbf{9.100} & 10.284 \\
			Recall@3: o.i. & 19.621 & 53.744 & \textbf{17.678} & 54.360 & 56.588 & 14.645 & 13.128 & \textbf{8.720} & 10.948 \\
 
			 \rowcolor{gray!30} Recall@5: m.i. & 27.915 & 67.109 & \textbf{26.066 }& 68.294 & 69.668 & 21.043 & 14.408 & \textbf{12.038} & 13.175 \\
			Recall@5: o.i. & 24.834 & 66.682 & \textbf{23.128} & 66.777 & 69.100 & 18.578 & 16.588 & \textbf{11.327} & 13.934 \\
 
			  \rowcolor{gray!30}Recall@10: m.i. & 35.261 & 74.739 & \textbf{33.175} & 76.161 & 77.014 & 28.057 & 20 & \textbf{17.441} & 17.820 \\
			Recall@10: o.i. & 32.322 & 75.308 & \textbf{30.521} & 74.455 & 77.393 & 24.502 & 23.128 & \textbf{16.351} & 18.957 \\
 
			  \rowcolor{gray!30}Recall@50: m.i. & 55.592 & 86.493 & \textbf{53.602} & 87.867 & 87.536 & 46.161 & 38.957 & \textbf{33.412} & 34.787 \\
			Recall@50: o.i. & 53.507 & 87.773 & \textbf{51.422} & 86.967 & 88.673 & 45.545 & 41.848 & \textbf{32.844} & 34.123 \\
 
			  \rowcolor{gray!30} Recall@100: m.i. & 65.308 & 91.232 & \textbf{63.697} & 91.469 & 91.754 & 55.498 & 48.389 & \textbf{42.749} & 44.455 \\
			Recall@100: o.i. & 63.744 & 91.896 & \textbf{61.327} & 91.611 & 91.943 & 55.024 & 51.090 & \textbf{42.559} & 45.166 \\
						\bottomrule
		\end{tabular}
	    \end{adjustbox}
	\end{center}
\caption{\label{st_vq_facenet_sphereface} Evaluation in the case of transfer to another embedding on the LFW dataset: PrivacyGAN (with VQGAN or StyleGAN) are optimised with FaceNet and tested with SphereFace. Generative methods do obtain better privacy results than Fawkes, except for the versions specifically created (weakened) to have privacy results similar to Fawkes (these versions are created for comparing image quality in Table \ref{humanstudytable} in a context with equal privacy performance).}
\end{table*}  

 \begin{table*}[htbp]
	\begin{center}
	    \begin{adjustbox}{max width=0.87\textwidth}

		\begin{tabular}{@{}l|rrrr|rHHHH@{}}
  \rowcolor{gray!30}
   \multicolumn{1}{c}{} & \multicolumn{4}{|c|}{PrivacyGAN} & \multicolumn{1}{c}{Pixel-based} & \multicolumn{4}{c}{}\\
			  & StyleGAN & StyleGAN & VQGAN & VQGAN & Fawkes & F + S & F + V & S + F & V + F \\
			   \rowcolor{gray!30}&&\_0.003\_500&&\_0.005\_128&&&&&\\
			\toprule
			Percentage  & \textbf{0.767} & 0.361 & 0.627 & 0.422 & 0.408 & 1.252 & \textbf{6.110} & 2.360 & 0.972 \\
 
			  \rowcolor{gray!30}Recall@1: m.i. & \textbf{20.758} & 24.028 & 24.028 & 24.265 & 24.882 & 19.763 & \textbf{0.616} & 15.782 & 19.905 \\
			Recall@1: o.i. & \textbf{21.611} & 25.972 & 22.701 & 25.545 & 25.403 & 22.654 & \textbf{11.090} & 18.815 & 21.280 \\
 
			  \rowcolor{gray!30}Recall@3: m.i. & \textbf{60.237} & 71.848 & 66.493 & 73.744 & 75.403 & 52.749 & \textbf{11.943} & 41.611 & 55.213 \\
			Recall@3: o.i. & \textbf{60.142} & 73.602 & 65.261 & 73.507 & 73.507 & 54.645 & \textbf{22.370} & 41.232 & 55.166 \\
 
			  \rowcolor{gray!30}Recall@5: m.i. & \textbf{78.294} & 96.967 & 86.209 & 98.389 & 98.768 & 69.005 & \textbf{21.754} & 55.545 & 73.791 \\
			Recall@5: o.i. & \textbf{76.777} & 97.156 & 83.744 & 98.436 & 98.863 & 67.536 & \textbf{27.820} & 51.232 & 69.573 \\
 
			  \rowcolor{gray!30}Recall@10: m.i. & \textbf{85.687} & 97.962 & 91.043 & 98.863 & 99.194 & 77.204 & \textbf{30.332} & 64.550 & 81.090 \\
			Recall@10: o.i. & \textbf{84.313} & 98.152 & 89.431 & 98.910 & 99.005 & 75.592 & \textbf{35.403} & 61.232 & 78.057 \\
 
			  \rowcolor{gray!30}Recall@50: m.i. & \textbf{93.223} & 99.005 & 95.545 & 99.005 & 99.194 & 88.246 & \textbf{50.948} & 78.768 & 90.444 \\
			Recall@50: o.i. & \textbf{92.512} & 98.957 & 95.450 & 99.052 & 99.194 & 87.630 & \textbf{55.498} & 77.678 & 89.621 \\
 
			  \rowcolor{gray!30}Recall@100: m.i. & \textbf{95.308} & 99.052 & 97.062 & 99.100 & 99.194 & 91.422 & \textbf{59.905} & 83.886 & 93.365 \\
			Recall@100: o.i. & \textbf{94.976} & 99.052 & 96.825 & 99.147 & 99.194 & 91.374 & \textbf{64.265} & 83.412 & 92.749 \\

		\bottomrule \end{tabular}
	    \end{adjustbox}
	\end{center}
\caption{ \label{st_vq_facenet_magface} Evaluation on the LFW dataset in the case of transfer to another embedding: PrivacyGAN (with VQGAN or StyleGAN) is optimised with FaceNet and tested with MagFace. Generative methods do obtain better privacy results than Fawkes, except for the versions specifically created (weakened) to have privacy results similar to Fawkes (these versions are created for comparing image quality in Table \ref{humanstudytable} in a context with equal privacy performance). {However, both Fawkes and generative methods optimised with one embedding do not transfer well to the novel embedding methods such as MagFace, while they transfer better to some other embedding methods such as SphereFace, as in Table \ref{st_vq_facenet_sphereface}.} }
\end{table*}
Examples of original images from the LFW dataset and their modifications obtained by our methods and by Fawkes are presented in Fig. \ref{image_facenet}. More examples are in the supplementary material.
\begin{figure*}[t] \begin{center}
\par
{\includegraphics[width=0.62\textwidth]{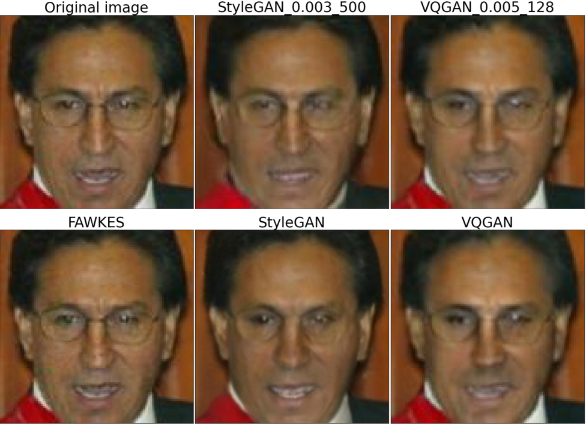} }%
\hfill
{\includegraphics[width=0.62\textwidth]{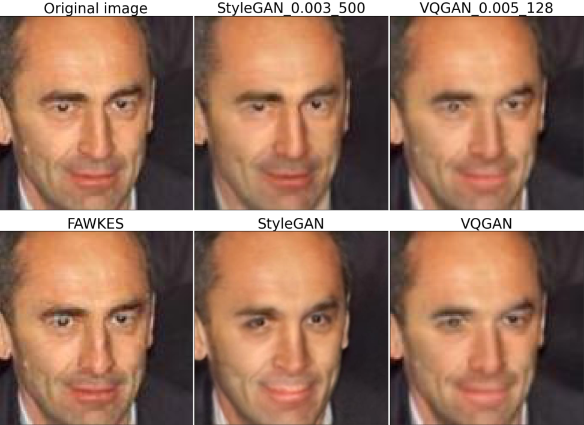} }%
\par
\end{center}
\caption{\label{image_facenet} Experiment 1: Examples of original images from the LFW dataset and their counterparts modified by different privacy methods: StyleGAN\_0.003\_500, VQGAN\_0.005\_128, StyleGAN, VQGAN (using FaceNet as an embedding method for optimization), and Fawkes. Here we can see that, while generative methods in general add more modification to an image than Fawkes, generative methods produce realistic images and do not add pixel noise. Study based on human ratings in Table \ref{humanstudytable}.}
\end{figure*}

\begin{table*}[htbp]
	\begin{center}
	    \begin{adjustbox}{max width=0.87\textwidth}
		
		\begin{tabular}{@{}cccccccHHHH@{}}
  \rowcolor{gray!30}
			  & StyleGAN & StyleGAN & StyleGAN & VQGAN & VQGAN & VQGAN \\
			   \rowcolor{gray!30}&&\_0.02\_500&\_0.02\_1000&&\_0.03\_512&\_0.04\_128\\
			\toprule
			Percentage  & \textbf{15.049} & 7.909 & 7.264 & 4.910 & 7.472 & 8.307 \\
 
			  \rowcolor{gray!30}Recall@1: m.i. & \textbf{0.095} & 0.806 & 0.900 & 0.521 & 0.284 & \textbf{0.095} \\
			Recall@1: o.i. & \textbf{3.555} & 8.768 & 10.521 & 11.185 & 6.588 & 6.919 \\
 
			  \rowcolor{gray!30}Recall@3: m.i. & \textbf{3.270} & 10.711 & 11.327 & 15.071 & 9.147 & 8.057 \\
			Recall@3: o.i. & \textbf{6.493} & 17.583 & 20.332 & 23.981 & 14.360 & 14.218 \\
 
			  \rowcolor{gray!30}Recall@5: m.i. & \textbf{5.118} & 16.919 & 19.479 & 26.682 & 17.583 & 14.834 \\
			Recall@5: o.i. & \textbf{8.863} & 21.943 & 25.071 & 30.664 & 19.147 & 17.867 \\
 
			  \rowcolor{gray!30}Recall@10: m.i. & \textbf{9.242} & 25.261 & 28.057 & 37.488 & 24.787 & 22.180 \\
			Recall@10: o.i. & \textbf{12.275} & 28.626 & 32.133 & 40 & 25.972 & 24.929 \\
 
			  \rowcolor{gray!30}Recall@50: m.i. & \textbf{23.649} & 44.550 & 46.919 & 57.678 & 44.692 & 42.085 \\
			Recall@50: o.i. & \textbf{26.114} & 48.436 & 50.521 & 60.806 & 46.967 & 44.028 \\
 
			  \rowcolor{gray!30}Recall@100: m.i. & \textbf{31.706} & 53.555 & 57.204 & 66.730 & 54.597 & 51.991 \\
			Recall@100: o.i. & \textbf{33.649} & 57.393 & 60.332 & 69.052 & 55.450 & 54.028 \\
   			\bottomrule
		\end{tabular}
		\end{adjustbox}
	\end{center}
\caption{Evaluation of various PrivacyGAN variants on the LFW dataset, case without transfer: PrivacyGAN (equipped with VQGAN and StyleGAN, including variants) are optimised with MagFace and MobileFaceNet and tested with MagFace. Lower recall means better privacy. Compared to Fawkes, results in Table \ref{st_vq_facenet_magface}: generative methods do get better privacy results. However, we did use MagFace in the algorithm, whereas Fawkes does not, hence the need for further validation (\ie, testing in the case of transfer to embeddings not used in the privacy algorithm), which is done, for example, in Table \ref{vq_magface_mobilefacenet_sphereface}. \label{lfw-5-fm-MagFace}
}
\end{table*}

Overall, from Tables \ref{facenet_facenet}, \ref{st_vq_facenet_sphereface} and \ref{st_vq_facenet_magface}, we note that with a standard set of parameters, VQGAN and StyleGAN are much better for privacy than Fawkes. In order to match Fawkes privacy results, we need to change the VQGAN and StyleGAN parameters tenfold. While these parameter changes decrease privacy significantly, generative methods still have a disruptive effect on image quality.

It is worth noting that optimising generative methods using a single embedding method is insufficient for adequate facial image privacy protection. In the case of transferring images to MagFace (Table \ref{st_vq_facenet_magface}), the correct identity for the modified image is often among the top 5 possibilities. Thus, in the next subsection, we use two different embedding methods in the optimisation process to generate private image versions.
Furthermore, we have observed that combining Fawkes poisoning with our proposed methods can be advantageous for facial image privacy protection. We expand on that in supplementary material.
\subsection{Experiment 2: Comparing StyleGAN and VQGAN Optimised with 2 Embedding Methods on the LFW Dataset}
We now compare standard versions of VQGAN and StyleGAN together with other specific versions: \begin{enumerate}
\item StyleGAN\_0.02\_500 and VQGAN\_0.04\_128; \item StyleGAN\_0.02\_1000 and VQGAN\_0.03\_512.
\end{enumerate}
These versions are proposed so that they have similar transfer recall scores in each pair.
{Specifically, average transfer recall scores for different methods are: $20.12\%$ for StyleGAN, $32.33\%$ for StyleGAN\_0.02\_500, $36.23\%$ for StyleGAN\_0.02\_1000, $42.41\%$ for VQGAN, $36.99\%$ for VQGAN\_0.03\_512 and $33.07\%$ for VQGAN\-\_0.04\-\_128. }
We create these specific versions of VQGAN and StyleGAN so that we can fairly compare the quality of the generated private images produced by the generative methods. We want to know which method produces the best image quality for a given threshold of our privacy metric.
An example of an evaluation result without transfer is presented in Table \ref{lfw-5-fm-MagFace} and for MobileFaceNet in the supplementary material.

\begin{table*}[htbp]
	\begin{center}
	    \begin{adjustbox}{max width=0.87\textwidth}
		
		\begin{tabular}{@{}lrrrrrHHHH@{}}
  \rowcolor{gray!30}
			  & StyleGAN & StyleGAN & StyleGAN  & VQGAN & VQGAN  & VQGAN  \\
			    \rowcolor{gray!30}			  &   &   \_0.02\_500 &  \_0.02\_1000 &   &  \_0.03\_512 &  \_0.04\_128 \\
			\toprule
			Percentage  & \textbf{12.273} & 8.157 & 7.715 & 6.211 & 7.387 & 8.283 \\
 
			  \rowcolor{gray!30}Recall@1: m.i. & \textbf{2.749} & 5.118 & 5.735 & 6.682 & 4.787 & 4.123 \\
			Recall@1: o.i. & \textbf{5.261} & 7.536 & 9.431 & 10.047 & 7.204 & 6.588 \\
 
			  \rowcolor{gray!30}Recall@3: m.i. & \textbf{6.256} & 11.801 & 12.701 & 14.408 & 12.512 & 9.858 \\
			Recall@3: o.i. & \textbf{9.573} & 14.218 & 18.199 & 17.630 & 14.408 & 12.607 \\
 
			  \rowcolor{gray!30}Recall@5: m.i. & \textbf{8.578} & 15.924 & 17.109 & 19.289 & 17.536 & 13.981 \\
			Recall@5: o.i. & \textbf{11.848} & 17.725 & 22.227 & 22.464 & 19.005 & 16.493 \\
 
			  \rowcolor{gray!30}Recall@10: m.i. & \textbf{13.033} & 20.711 & 22.891 & 26.398 & 24.360 & 19.763 \\
			Recall@10: o.i. & \textbf{15.877} & 24.218 & 27.441 & 29.858 & 26.303 & 22.227 \\
 
			  \rowcolor{gray!30}Recall@50: m.i. & \textbf{26.209} & 39.858 & 41.943 & 47.441 & 42.749 & 38.341 \\
			Recall@50: o.i. & \textbf{30.379} & 43.554 & 45.403 & 51.659 & 46.161 & 40.284 \\
 
			  \rowcolor{gray!30}Recall@100: m.i. & \textbf{35.024} & 49.336 & 50.995 & 57.820 & 52.749 & 49.194 \\
			Recall@100: o.i. & \textbf{39.716} & 52.654 & 55.024 & 61.754 & 56.256 & 52.559 \\
			\bottomrule
		\end{tabular}
    \end{adjustbox}
	\end{center}
\caption{Evaluation of various PrivacyGAN variants in the case of transfer to another embedding on the LFW dataset: PrivacyGAN equipped with VQGAN or StyleGAN is optimised with MagFace and MobileFaceNet and tested with SphereFace.\label{vq_magface_mobilefacenet_sphereface} In this case, generative methods optimised with two embeddings obtain better results than generative methods optimised with only one embedding method in Table \ref{st_vq_facenet_sphereface}. }
\end{table*}

We also study the transfer to different embedding methods. One of the results of this study (for the embedding method SphereFace) is presented in Table \ref{vq_magface_mobilefacenet_sphereface}. Transfer results for other embedding methods can be found in the supplementary material.

\begin{figure*}[htbp] \begin{center}
\par
{\includegraphics[width=0.62\textwidth]{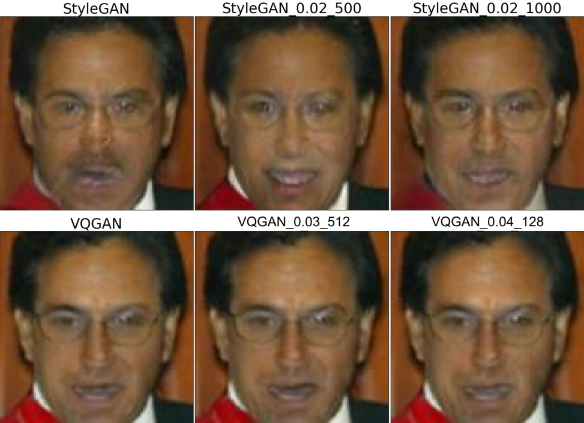} }%
\par
\caption{\label{image_magface_MobileFaceNet}
Experiment 2: Examples of images from the LFW dataset modified by different privacy methods: StyleGAN, StyleGAN\_0.02\_500, StyleGAN\_0.02\_1000, VQGAN, VQGAN\_0.03\_512, VQGAN\_0.04\_128 with embedding methods MagFace and MobileFaceNet. {Original and modified Fawkes image versions can be seen in Fig. \ref{image_facenet}. {These images have different privacy levels and different qualities: }the human rating experiment performs comparisons between images produced by methods with similar recall, \ie, similar privacy results.} }
\end{center}
\end{figure*} From the table \ref{vq_magface_mobilefacenet_sphereface} compared to \ref{st_vq_facenet_sphereface} we can see that, in general, generative methods optimised with two embeddings transfer better to other embedding methods than generative methods optimised with only one embedding. For instance, for the unused in an optimisation process embedding SphereFace, the percentage score for StyleGAN with standard parameters optimised with $2$ different embedding methods is more than $15$ while it was just around $5$ for one embedding method. Examples of images {produced} by the methods of experiment 2 are presented in Fig. \ref{image_magface_MobileFaceNet}. More of the examples can be found in the supplementary material.

\begin{table*}[htbp]
	\begin{center}
	\begin{adjustbox}{max width=0.87\textwidth}
		
		\begin{tabular}{@{}ccHHHHcHHHHHHHHHccHHcHc@{}}
  \rowcolor{gray!30}
   \multicolumn{1}{c}{} & \multicolumn{16}{c}{PrivacyGAN} & \multicolumn{1}{c}{Pixel-based} & \multicolumn{2}{c}{} & \multicolumn{1}{c}{PrivacyGAN} & \multicolumn{1}{c}{} & \multicolumn{1}{c}{Adversarial} \\

			 & VQGAN & VQGAN & VQGAN\_0.007\_128 & VQGAN & VQGAN & VQGAN & VQGAN & VQGAN & VQGAN & VQGAN & VQGAN & VQGAN & VQGAN & VQGAN & VQGAN& VQGAN & Fawkes & Fawkes & Fawkes & StyleGAN & StyleGAN & AMT-GAN \\
		  \rowcolor{gray!30}	&  \_0.003\_128 & \_0.005\_128 & \_0.007\_128 & \_0.01\_128 & \_0.02\_128 &  & \_0.03\_256 & \_0.03\_512 & \_0.03\_1024 & \_0.03\_2048 & \_0.04\_128 & \_0.04\_256 & \_0.04\_512 & \_0.04\_1024 & \_0.04\_2048 & \_0.04\_4096 &  & \_{high4} & \_{high5} & \_0.02\_1000 & \_0.01\_500 & \\
  			\toprule
			Percentage & 9.424 & 9.782 & 10.114 & 10.635 & 12.007 & 13.399 & 13.896 & 14.291 & 14.565 & 14.653 & 14.498 & 15.258 & 15.633 & 15.904 & 16.130 & 16.379 & 7.519 & 10.676 & 11.439 & \textbf{17.024} & 13.030 & 14.067 \\
		  \rowcolor{gray!30}	Recall@1: m.i. & 17.854 & 16.871 & 15.727 & 14.162 & 11.474 & 9.529 & 8.646 & 8.365 & 8.124 & 7.703 & 7.823 & 7.623 & 6.820 & 6.841 & 6.520 & 5.998 & 23.952 & 16.690 & 15.266 & \textbf{5.416} & 9.910 & 9.328 \\
			Recall@1: o.i. & 18.034 & 17.071 & 16.830 & 15.928 & 12.357 & 11.214 & 10.632 & 10.030 & 10.030 & 8.967 & 9.248 & 9.007 & 8.826 & 7.382 & 8.144 & \textbf{6.800} & 24.092 & 16.229 & 14.463 & {6.841} & 11.675  & 8.445 \\
		  \rowcolor{gray!30}	Recall@3: m.i. & 40.702 & 38.616 & 36.369 & 33.681 & 25.737 & 21.364 & 20.020 & 18.596 & 18.114 & 17.573 & 17.833 & 16.489 & 15.386 & 14.925 & 14.704 & 13.561 & 52.979 & 32.377 & 28.887 & \textbf{12.197} & 22.327 & 19.980 \\
			Recall@3: o.i. & 41.765 & 39.278 & 37.011 & 34.263 & 27.081 & 23.149 & 22.006 & 21.103 & 19.980 & 19.699 & 19.218 & 18.576 & 17.111 & 16.229 & 16.369 & 14.483 & 53.420 & 30.552 & 27.803 & \textbf{14.443} & 24.373 & 17.593 \\
		  \rowcolor{gray!30}	Recall@5: m.i. & 48.245 & 45.737 & 43.470 & 39.819 & 31.133 & 26.439 & 24.273 & 23.029 & 21.886 & 21.705 & 21.886 & 20.983 & 19.478 & 18.455 & 18.475 & 17.051 & 61.244 & 37.753 & 34.223 & \textbf{15.727} & 26.941 & 24.293 \\
			Recall@5: o.i. & 49.629 & 46.680 & 44.173 & 40.943 & 32.297 & 27.924 & 26.560 & 24.875 & 24.132 & 23.651 & 23.290 & 22.708 & 20.943 & 19.759 & 20.020 & \textbf{17.994} & 61.224 & 35.466 & 32.538 & {18.134} & 29.809 & 21.344 \\
		  \rowcolor{gray!30}	Recall@10: m.i. & 53.220 & 50.772 & 48.425 & 45.396 & 36.369 & 31.515 & 29.168 & 28.305 & 27.081 & 26.419 & 27.302 & 25.597 & 24.313 & 23.069 &\textbf{ 22.969} & 21.143 & 65.055 & 42.909 & 39.037 & \textbf{20.100} & 32.959  & 29.228 \\
			Recall@10: o.i. & 54.945 & 52.036 & 49.970 & 46.239 & 37.854 & 33.621 & 31.735 & 29.448 & 29.428 & 28.987 & 28.746 & 27.723 & 25.757 & \textbf{24.514} & 24.975 & \textbf{22.768 }& 65.135 & 40.562 & 37.051 & 23.531 & 36.008 & 25.436 \\
		  \rowcolor{gray!30}	Recall@50: m.i. & 64.253 & 61.565 & 60.461 & 57.332 & 50.491 & 44.835 & 43.370 & 41.123 & 40.562 & 39.639 & 40.321 & 38.957 & 36.670 & 35.988 & 35.326 & 34.343 & 72.979 & 54.985 & 51.254 & \textbf{32.618} & 47.161 & 42.086 \\
			Recall@50: o.i. & 65.537 & 63.069 & 61.685 & 58.877 & 51.836 & 47.442 & 45.055 & 43.671 & 43.631 & 41.946 & 42.367 & 41.003 & 38.736 & 37.733 & 37.914 & \textbf{36.068 }& 72.738 & 52.638 & 48.506 & {36.911} & 50.652 & 36.409 \\
		  \rowcolor{gray!30}	Recall@100: m.i. & 68.726 & 67.101 & 65.316 & 62.588 & 56.469 & 51.675 & 50.451 & 48.506 & 47.763 & 46.800 & 47.583 & 46.259 & 44.032 & 43.009 & 42.528 & 41.484 & 76.108 & 60.100 & 56.971 &\textbf{ 39.478 } & 53.882  & 49.509 \\
			Recall@100: o.i. & 70.030 & 68.044 & 67.041 & 64.173 & 57.813 & 54.363 & 52.156 & 50.772 & 50.812 & 48.987 & 49.168 & 48.225 & 46.299 & {45.115} & 45.055 & {44.152 }& 75.928 & 58.094 & 54.403 & 45.176 & 57.212 & \textbf{42.467} \\
   			\bottomrule
		\end{tabular} \end{adjustbox}
	\end{center}
\caption{ \label{transfer_CC_sph} Evaluation in the case of a transfer to another embedding on the CC dataset: VQGAN and StyleGAN are optimised with MagFace and MobileFaceNet and tested with SphereFace. PrivacyGAN basically outperforms Fawkes while the comparison with AMT-GAN (which could be used on top of our method) depends on criteria and parameters.}
\end{table*}
\subsection{Experiment 3: Comparing StyleGAN, VQGAN, and Fawkes on the CC dataset \label{CC_experiment}}

Here we choose the specific versions of VQGAN, StyleGAN, and Fawkes that have similar transfer recall scores in each group for the dataset CC: 

\begin{enumerate}
\item VQGAN\_0.003\_128 and Fawkes;
\item Style\-GAN\-\_0.02\_1000 and VQGAN\_0.04\_4096.
\end{enumerate}
In addition, we have compared our results to those obtained with AMT-GAN. However, it is important to note that a direct comparison between our proposed methods and AMT-GAN is not possible, as AMT-GAN is unable to transfer makeup to faces that were not detected. Therefore, in cases where faces were not detected, we had to replace them with the original images, which may affect the comparability of the results.
Transfer recalls for the proposed methods are the following: 
AMT-GAN: $49.57\%$,
Fawkes: $79.21\%$,
StyleGAN\_0.02\_1000: $24.71\%$,
VQGAN\_0.003\_128: $74.76\%$, VQGAN\_0.04\_4096: $26.09\%$, VQGAN: $40.45\%$.
We compare how well proposed generative and pixel-based approaches protect privacy against different embedding methods. One of the results of this study (for the embedding method SphereFace) is presented in Table \ref{transfer_CC_sph}. Transfer results for other embedding methods can be found in the supplementary material. Examples of images {produced} by the methods of experiment $3$ are presented in {Fig.} \ref{CC}. More examples can be found in the supplementary material.

\begin{figure*}[htbp] 
\begin{center}
\par
{\includegraphics[width=0.62\textwidth]{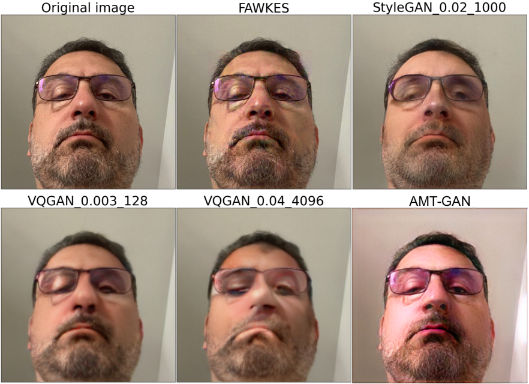} }%
\par


\caption{\label{CC}{Experiment 3: {Examples of images of volunteers modified by various privacy methods,} including AMT-GAN, Fawkes and StyleGAN\_0.02\_1000, VQGAN\_0.003\_128, VQGAN\_0.04\_4096 optimised with embedding methods MagFace and MobileFaceNet. {The different methods and parametrizations lead to different image quality/privacy results; the human rating experiments will compare the quality for methods with similar recall.}}}
\end{center}
\end{figure*} 

Using table \ref{transfer_CC_sph}, we can conclude that, despite using the CC dataset instead of LFW, generative methods prove to be effective for privacy preservation and tend to outperform both the pixel-based method Fawkes and the generative makeup transfer method AMT-GAN.
 
\subsection{Human preferences for similar transfer recall}\label{sectionhumanstudy}

{In {Figs} \ref{CC} and \ref{image_magface_MobileFaceNet}, we can see that, in some cases, {modifying} the number of iterations in optimisation and the coefficient {$K$ affects} the quality of an image {and} its privacy protection. {Therefore, to} evaluate the modified image quality for different privacy methods with similar transfer recall, we conducted a human preference study.
Three human raters were presented with $40$ pairs of images generated by the methods discussed in section \ref{CC_experiment}. Given two images generated by two different methods, the human rater could choose ``I prefer the left one as an avatar,'' ``I prefer the right one as an avatar,'' or ``No preference.''
To provide context, the human raters involved in the experiment were not paid and were not authors of the current paper. They were selected using the snowball principle, and their task was to assess the quality and similarity of the modified image to its original version. Without this assessment, we could end up with a black square instead of a privacy-protected image. The human raters did not have a degree in computer science and were not informed that the experiment related to privacy. However, the instructions provided to them, which included presenting the original image at the centre and emphasising that all images were reasonably close to it, as well as providing examples of potential use cases such as social networks, news articles, and dating websites, made it clear that assessing image similarity was an integral part of the task.
The human-assessment results are presented in Table \ref{humanstudytable}.}

 \begin{table*}[htbp]
     \begin{center}
     \begin{adjustbox}{max width=0.87\textwidth}
     
		\begin{tabular}{@{}lrrrrrHHHH@{}}
  \rowcolor{gray!30}
   & & & Human preference: \\
\rowcolor{gray!30} Transfer recall & Avatar & Avatar & success rate\\
  \rowcolor{gray!30}             & method 1 & method 2 & of 1 vs 2 \\
        \hline
 \rowcolor{gray!15}       \multicolumn{4}{|c|}{{High privacy (low recall), LFW dataset}}\\
        \hline
 32.6\% & (StyleGAN\_0.02\_500,    &{\bf{(VQGAN\_128\_0.04,}}  &  43.75$\pm$5\%\\
   & MagFace + MobileFaceNet) &{\bf{MagFace + MobileFaceNet) }} & \\
   
   36.7\% & (StyleGAN\_0.02 1000,   & {\bf{(VQGAN\_0.03\_512,}}  &  35.37\% $\pm$ 5\% \\
        & MagFace + MobileFaceNet) & {\bf{MagFace + MobileFaceNet)}}& \\
        \hline
     \rowcolor{gray!15}   \multicolumn{4}{|c|}{Low privacy (high recall), LFW dataset}\\
        \hline
        90\% & {\bf{(StyleGAN\_0.003\_500}} & Fawkes & 87.2\% $\pm$ 2\% \\
        & {\bf{FaceNet)}} & &\\
        \hline
      \rowcolor{gray!15}  \multicolumn{4}{|c|}{{High privacy, (low recall), CC dataset}}\\
        \hline 
        26.09\%  -  24.71\%    & {\bf{(VQGAN\_0.04\_4096,}}
     & (StyleGAN\_0.02\_1000,  & 55.2  \% $\pm$ 3.59\%\\
        
        & {\bf{MagFace + MobileFaceNet)}} & MagFace + MobileFaceNet)& \\
        \hline
       \rowcolor{gray!15} \multicolumn{4}{|c|}{{Low privacy, (high recall), CC dataset}}\\
        \hline 
        74.76\%  -  79.21\%  & {\bf{(VQGAN\_0.003\_128)}} & Fawkes & 51.5 \% $\pm$ 3.06\%\\
        & {\bf{MagFace + MobileFaceNet)}} & &\\
     \bottomrule \end{tabular}
     \end{adjustbox}
     \end{center}
\caption{\label{humanstudytable} We modify the strength of different privacy-protection-algorithm perturbations until we get to similar target recall levels. 
We compare the quality of images, for each recall level.
{{Text in bold font}} refers to human preference, for each recall level (see rightmost column).}
\end{table*}

We compared $5$ different pairs of privacy-preserving methods with similar target recall values. In the low privacy (high transfer recall) setting for both LFW and CC datasets, we were able to compare{, in terms of quality, the Fawkes} method with generative methods specifically modified to match Fawkes transfer recall values. When we choose FaceNet as an embedding for generative methods (StyleGAN) optimisation, {the quality} of the images generated by StyleGAN appears to be worse than that of Fawkes. However, when we use MagFace and MobileFaceNet as embeddings for generative methods optimisation, we obtain similar image quality results for VQGAN and Fawkes ($51.5 \% \pm 3.06 \%$), while VQGAN has a better transfer recall ($74.76 \%$ compared to $79.21 \%$).
In the high privacy (low transfer recall) setting for both the LFW and CC datasets, we were able to compare different modifications of the generative methods (VQGAN and StyleGAN) with similar transfer recall. In every case, it appears that human raters preferred VQGAN-generated images over StyleGAN-generated images.

\subsection{Human Identification of Same-Person Images}
\label{sectionhumanevalsimilarity}

In this section, our objective is to evaluate the effectiveness of various privacy preservation methods for their applicability on social media platforms and to determine the extent to which people can identify the person after privacy-preservation modifications by different methods, namely: VQGAN\_0.005\_128, AMT-GAN, and Fawkes, and the anonymization method Deep Privacy 2 \ref{hukkelas23DP2}.

The experiment is structured as follows: we begin with the original image, referred to as the ``original'' in the filename. We then examine privacy-preserved versions of different images of the same individual, followed by random images of different people. The central question for each image in a given set is, ``Is this the same person as in the original?''.

\begin{figure*}[htbp]
    \centering
    \includegraphics[width=0.95\textwidth]{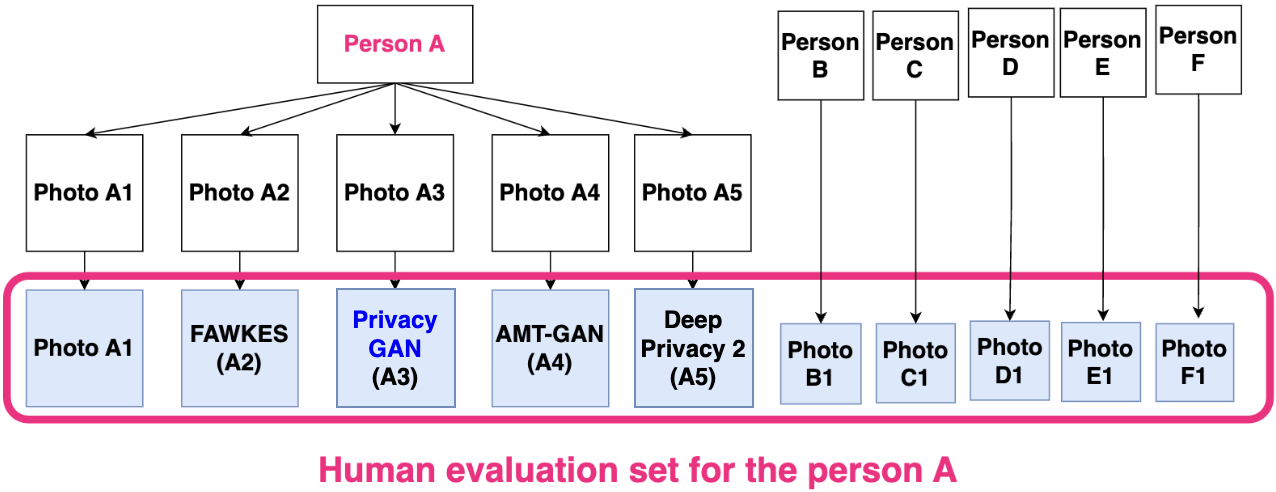}
    \caption{Schema of the human identification experiment setup. The experiment involves comparing the original image with privacy-preserved versions of images of the same person and random images of different people to determine whether privacy-preservation methods could preserve the utility of the images modified by them: we expect human raters to recognize the original face for privacy-preserving methods, and not for anonymization methods.}
    \label{fig:human_exp_schema}
\end{figure*}

To execute this experiment, we established a setup illustrated in Figure~\ref{fig:human_exp_schema}. This schema visually represents the process, illustrating the different image types involved in the human identification experiment.

We recruited $5$ human evaluators, aged from $15$ to $43$ years, to assess pairs of images, comprising the original image and 
constructed using privacy-preserving methods for the same individual,
as outlined in the schema in Figure~\ref{fig:human_exp_schema}. The results of the human study are presented in Table~\ref{hi}.

\begin{table*}[htbp]
\centering

\begin{tabular}{lcc}
\toprule
\rowcolor{gray!30}Model & Accuracy & 99\% confidence interval \\
\rowcolor{gray!30} & for humans & \\
\midrule
PrivacyGAN using VQGAN\_0.005\_128 & 0.796 & $\pm$ 0.04 \\
\rowcolor{gray!30}AMT-GAN & 0.829 & $\pm$ 0.038 \\
Fawkes & 0.842 & $\pm$ 0.037 \\
\rowcolor{gray!30} Deep Privacy 2 & 0.187 & $\pm$ 0.039 \\
\bottomrule 
\end{tabular}

\caption{Human face identification for various privacy preservation models. The table demonstrates that, while our method VQGAN\_0.005\_128, together with AMT-GAN and Fawkes, generate recognisable images, the anonymization method Deep Privacy 2 often produces images that cannot be recognised as the same person. The purpose and limitations of this experiment are further discussed in the text. \label{hi}}
\end{table*}

From the findings presented in Table~\ref{hi}, we deduce that VQGAN\_0.005\_128, AMT-GAN, and Fawkes generate images that are similarly identifiable by human evaluators, with approximately $80\%$ of images generated by these methods being successfully recognised as the same as the original. Conversely, the anonymization method Deep Privacy 2 frequently produces images that cannot be identified as those of the same individual. This further substantiates that while anonymization methods such as Deep Privacy 2 preserve certain attributes of images, they may fail to preserve their utility. In contrast, our method, along with other utility-focused methods, while not providing absolute protection against facial recognition, effectively safeguards human facial images against prevalent face recognition techniques and maintains image utility for social media use.

This experiment was designed specifically to showcase that the recognisability of our method, along with AMT-GAN and Fawkes, is higher than that of anonymisation methods such as Deep Privacy 2. This explains our focus on testing just one version of PrivacyGAN. In future experiments, we intend to incorporate multiple versions of PrivacyGAN, and to involve a larger pool of human raters. This approach aims to determine which versions of PrivacyGAN perform best in terms of privacy/recall balance.

\section{Limitations and Future Work}

While generative methods are effective in safeguarding image privacy against various embedding methods, they cannot be compared to anonymization techniques. As \ref{radiya2021data} argues, it is always possible for new recognition attacks to be effective against provided data poisoning methods. However, our approach is different from anonymization,
as we do not aim to provide a privacy guarantee against future attacks. Instead, our objective is to protect users against stalkers and unauthorized identification
using current state-of-the-art recognition methods, while still enabling them to share their photos online, and be recognised by their family and friends. In the future, we would also like to use face enhancement as a tool for or against privacy methods, and check how much PrivacyGAN can be combined with AMT-GAN.

\section{Conclusions}
Our contributions are

\begin{enumerate}
\item A new approach to privacy based on inspirational generation, namely PrivacyGAN, using generative models for generating faces close to a given target. This method is orthogonal to the principles of AMT-GAN, so that our method could be used as a first step before AMT-GAN.

\item A comparison between these methods and traditional pixel-based methods, including transfer to unknown embeddings (a.k.a. robustness to unknown embeddings used for identifying people) and human raters for validating image quality.

\item A new privacy evaluation method based on the percentage of dataset images that are closer in an embedding space to a modified ``private'' image than to an original image.

\item A new dataset extracted from CC (more details are provided in the supplementary material). At the end, we recommend generative methods (Alg.~\ref{alggen}), with several embeddings so that robustness and transfer to new methods are properly tested.
\end{enumerate}

According to the human ratings study, Fawkes might be better than StyleGAN for generating high-quality images in the category ``low privacy'' ({recall rate of 90\%}) {on LFW.} However, VQGAN and Fawkes have similar results in a low-privacy (74.76-79.21\% as a transfer recall) setting, while VQGAN provides better privacy protection.
Among the proposed generative methods,
VQGAN is better than StyleGAN overall in terms of quality for a given privacy threshold (see Table \ref{humanstudytable}). By the human identification study (section \ref{sectionhumanevalsimilarity}) we further show that, in contrast to anonymization methods, our method, along with other utility-focused methods, effectively safeguards facial image privacy against prevalent face recognition techniques while maintaining image utility for social media use.

In comparison to AMT-GAN, our method demonstrates superior privacy outcomes depending on the parameter settings, although it doesn't necessarily enhance human recognizability. While AMT-GAN excels in scenarios where recognizability is high and privacy is low, PrivacyGAN offers a broader spectrum of applications, particularly in cases where definitive facial detection is challenging, or where makeup contradicts the user's personal beliefs.

\FloatBarrier

\section*{References}

\begin{enumerate}[label={[\arabic*]}]

\item \label{barattin2023attribute}
Simone Barattin, Christos Tzelepis, Ioannis Patras, and Nicu Sebe.
\emph{Attribute-preserving face dataset anonymization via latent code optimization}.
In Proceedings of the IEEE/CVF Conference on Computer Vision and Pattern Recognition, pages 8001–8010, 2023.
  
\item \label{chen2018mobilefacenets}
Sheng Chen, Yang Liu, Xiang Gao, and Zhen Han.
\emph{Mobilefacenets: Efficient cnns for accurate real-time face verification on mobile devices}.
In Chinese Conference on Biometric Recognition, pages 428–438. Springer, 2018.

\item \label{cherepanova2021Lowkey}
Valeriia Cherepanova, Micah Goldblum, Harrison Foley, Shiyuan Duan, John Dickerson, Gavin Taylor, and Tom Goldstein.
\emph{Lowkey: Leveraging adversarial attacks to protect social media users from facial recognition}.
arXiv preprint arXiv:2101.07922, 2021.

\item \label{dabouei2019fast}
Ali Dabouei, Sobhan Soleymani, Jeremy Dawson, and Nasser Nasrabadi.
\emph{Fast geometrically-perturbed adversarial faces}.
In 2019 IEEE Winter Conference on Applications of Computer Vision (WACV), pages 1979–1988. IEEE, 2019.

\item \label{deng2019ArcFace}
Jiankang Deng, Jia Guo, Niannan Xue, and Stefanos Zafeiriou.
\emph{Arcface: Additive angular margin loss for deep face recognition}.
In Proceedings of the IEEE/CVF conference on computer vision and pattern recognition, pages 4690–4699, 2019.

\item \label{esser2021taming}
Patrick Esser, Robin Rombach, and Bjorn Ommer.
\emph{Taming transformers for high-resolution image synthesis}.
In Proceedings of the IEEE/CVF conference on computer vision and pattern recognition, pages 12873–12883, 2021.

\item \label{Gafni:2019}
Oran Gafni, Lior Wolf, and Yaniv Taigman.
\emph{Live face de-identification in video}.
In Proceedings of the IEEE/CVF International Conference on Computer Vision, pages 9378–9387, 2019.

\item \label{harwell2022facial}
Drew Harwell.
\emph{This facial recognition website can turn anyone into a cop—or a stalker}.
In Ethics of Data and Analytics, pages 63–67. Auerbach Publications, 2022.

\item \label{hazirbas2021towards}
Caner Hazirbas, Joanna Bitton, Brian Dolhansky, Jacqueline Pan, Albert Gordo, and Cristian Canton Ferrer.
\emph{Towards measuring fairness in ai: the casual conversations dataset}.
IEEE Transactions on Biometrics, Behavior, and Identity Science, 2021.

\item \label{he2016deep}
Kaiming He, Xiangyu Zhang, Shaoqing Ren, and Jian Sun.
\emph{Deep residual learning for image recognition}.
In Proceedings of the IEEE conference on computer vision and pattern recognition, pages 770–778, 2016.

\item \label{hellmann2023ganonymization}
Fabio Hellmann, Silvan Mertes, Mohamed Benouis, Alexander Hustinx, Tzung-Chien Hsieh, Cristina Conati, Peter Krawitz, and Elisabeth André.
\emph{Ganonymization: A gan-based face anonymization framework for preserving emotional expressions}.
arXiv preprint arXiv:2305.02143, 2023.

\item \label{hu2022protecting}
Shengshan Hu, Xiaogeng Liu, Yechao Zhang, Minghui Li, Leo Yu Zhang, Hai Jin, and Libing Wu.
\emph{Protecting facial privacy: generating adversarial identity masks via style-robust makeup transfer}.
In Proceedings of the IEEE/CVF Conference on Computer Vision and Pattern Recognition, pages 15014–15023, 2022.

\item \label{huang2008labeled}
Gary B Huang, Marwan Mattar, Tamara Berg, and Eric Learned-Miller.
\emph{Labeled faces in the wild: A database for studying face recognition in unconstrained environments}.
In Workshop on faces in’Real-Life’Images: detection, alignment, and recognition, 2008.

\item \label{hukkelaas2019deepprivacy}
Håkon Hukkelås, Rudolf Mester, and Frank Lindseth.
\emph{Deepprivacy: A generative adversarial network for face anonymization}.
In International symposium on visual computing, pages 565–578. Springer, 2019.

\item \label{hukkelas23DP2}
Håkon Hukkelås and Frank Lindseth.
\emph{Deepprivacy2: Towards realistic full-body anonymization}.
In 2023 IEEE/CVF Winter Conference on Applications of Computer Vision (WACV), pages 1329–1338, 2023.

\item \label{karras2019style}
Tero Karras, Samuli Laine, and Timo Aila.
\emph{A style-based generator architecture for generative adversarial networks}.
In Proceedings of the IEEE/CVF conference on computer vision and pattern recognition, pages 4401–4410, 2019.

\item \label{Kim:2019}
Taehoon Kim and Jihoon Yang.
\emph{Latent-space-level image anonymization with adversarial protector networks}.
IEEE Access, 7:84992–84999, 2019.

\item \label{liu2017SphereFace}
Weiyang Liu, Yandong Wen, Zhiding Yu, Ming Li, Bhiksha Raj, and Le Song.
\emph{Sphereface: Deep hypersphere embedding for face recognition}.
In Proceedings of the IEEE conference on computer vision and pattern recognition, pages 212–220, 2017.

\item \label{meng2021MagFace}
Qiang Meng, Shichao Zhao, Zhida Huang, and Feng Zhou.
\emph{Magface: A universal representation for face recognition and quality assessment}.
In Proceedings of the IEEE/CVF Conference on Computer Vision and Pattern Recognition, pages 14225–14234, 2021.

\item \label{qiu2022novel}
Yuying Qiu, Zhiyi Niu, Biao Song, Tinghuai Ma, Abdullah Al-Dhelaan, and Mohammed Al-Dhelaan.
\emph{A novel generative model for face privacy protection in video surveillance with utility maintenance}.
Applied Sciences, 12(14):6962, 2022.

\item \label{radiya2021data}
Evani Radiya-Dixit and Florian Tramèr.
\emph{Data poisoning won’t save you from facial recognition}.
arXiv preprint arXiv:2106.14851, 2021.

\item \label{sample2019facial}
Ian Sample.
\emph{What is facial recognition-and how sinister is it}.
The Guardian, 29, 2019.

\item \label{schroff2015facenet}
Florian Schroff, Dmitry Kalenichenko, and James Philbin.
\emph{Facenet: A unified embedding for face recognition and clustering}.
In Proceedings of the IEEE conference on computer vision and pattern recognition, pages 815–823, 2015.

\item \label{shan2020Fawkes}
Shawn Shan, Emily Wenger, Jiayun Zhang, Huiying Li, Haitao Zheng, and Ben Y Zhao.
\emph{Fawkes: Protecting privacy against unauthorized deep learning models}.
In 29th USENIX security symposium (USENIX Security 20), pages 1589–1604, 2020.

\item \label{sharif2016accessorize}
Mahmood Sharif, Sruti Bhagavatula, Lujo Bauer, and Michael K Reiter.
\emph{Accessorize to a crime: Real and stealthy attacks on state-of-the-art face recognition}.
In Proceedings of the 2016 acm sigsac conference on computer and communications security, pages 1528–1540, 2016.

\item \label{wang2021facex}
Jun Wang, Yinglu Liu, Yibo Hu, Hailin Shi, and Tao Mei.
\emph{Facex-zoo: A pytorch toolbox for face recognition}.
In Proceedings of the 29th ACM International Conference on Multimedia, pages 3779–3782, 2021.

\item \label{wang2021variational}
Kuan-Chieh Wang, Yan Fu, Ke Li, Ashish Khisti, Richard Zemel, and Alireza Makhzani.
\emph{Variational model inversion attacks}.
Advances in Neural Information Processing Systems, 34:9706–9719, 2021.

\item \label{weise2009face}
Thibaut Weise, Hao Li, Luc Van Gool, and Mark Pauly.
\emph{Face/off: Live facial puppetry}.
In Proceedings of the 2009 ACM SIGGRAPH/Eurographics Symposium on Computer animation, pages 7–16, 2009.

\item \label{wenger2023sok}
Emily Wenger, Shawn Shan, Haitao Zheng, and Ben Y Zhao.
\emph{Sok: Anti-facial recognition technology}.
In 2023 IEEE Symposium on Security and Privacy (SP), pages 864–881. IEEE, 2023.

\item \label{fawkesgithub} Shawn Shan, Emily Wenger, Jiayun Zhang, Huiying Li, Haitao Zheng, and Ben Y
Zhao. Fawkes: Protecting privacy against unauthorized deep learning models. In
FAWKES \\\text{github: https://github.com/Shawn-Shan/fawkes, 2020}

\item \label{lpips}
Richard Zhang, Phillip Isola, Alexei A Efros, Eli Shechtman, and Oliver Wang.
\emph{The unreasonable effectiveness of deep features as a perceptual metric}.
In Proceedings of the IEEE conference on computer vision and pattern recognition, pages 586–595, 2018.

\end{enumerate}

\newpage
\appendix
\section{Appendix}

Facial image privacy protection is a multi-objective problem combining image quality preservation and privacy robustness against various image recognition systems:
\begin{itemize}
\item In Section \ref{amtsect}, we provide the description of generative makeup transfer method AMT-GAN
\item In Sections \ref{s1b}-\ref{s2}, we present quantitative results (table of recall/percentage, showing privacy performance).
\item Then, in Sections \ref{s3}-\ref{s4}, we present images, showing the image quality.
\end{itemize}
A human evaluation study that combines both privacy and image quality aspects is available in the main paper (Table~\ref{humanstudytable}, Table~\ref{hi}).

\subsection{Privacy algorithms: generative makeup transfer method AMT-GAN \label{amtsect}}

In this paper, we compare the results of our method PrivacyGAN based on generative techniques such as StyleGAN \ref{karras2019style} and VQGAN\ref{esser2021taming} to a method for generative makeup transfer known as AMT-GAN~\ref{hu2022protecting}. The objective of AMT-GAN is to produce adversarial images that incorporate the makeup style of reference images. Although AMT-GAN introduces more alterations to the original image, it confines these modifications to the makeup application areas, thus resulting in visually natural images, as demonstrated by the FID results. The authors of the method employ LPIPS~\ref{lpips} loss to retain image similarity to the original, which we also utilise in our paper.

In the main body of the paper, we mention that while photographs of individuals wearing makeup may look appropriate and natural to some people, there are certain drawbacks to this approach. Firstly, publishing photos with makeup may be deemed unsuitable for certain groups of individuals. Secondly, if the face in the photograph is not clearly discernible, AMT-GAN may not recognise it and may not generate a private version of the image, unlike Fawkes~\ref{shan2020Fawkes} and our method.

\subsection{Experiment 1: additional and extended tables of results}\label{s1b}

In this subsection, we present transfer results for Experiment 1. All the tables here are similar to the Table \ref{st_vq_facenet_sphereface} except for the different choice of transfer embeddings used in recognition.

To be precise, we evaluate the transfer results of PrivacyGAN equipped with generative methods VQGAN and StyleGAN optimised with FaceNet\ref{schroff2015facenet} embedding and compare them to the transfer results of Fawkes. The criterion is the transfer to other embeddings than FaceNet, namely ArcFace\ref{deng2019ArcFace}, MobileFaceNet\ref{chen2018mobilefacenets}, and Resnet\_152\ref{he2016deep}.

We also present the results of Fawkes combination with generative methods. We note that combining Fawkes poisoning with our methods can be beneficial for facial image privacy protection.

Table \ref{st_vq_facenet_arcface} presents results of the transfer to ArcFace embedding; Table \ref{st_vq_facenet_mobilefacenet} presents results of the transfer to MobileFaceNet embedding; and Table \ref{vq_st_facenet_resnet}
presents the results of transfer to Resnet\_152 embedding method.

From the results of this experiment, we conclude that generative methods optimised with one single embedding do not provide strong privacy protection, but their results are still better than the results of Fawkes.

\subsection{Experiment 2 (comparing PrivacyGAN equipped with StyleGAN and PrivacyGAN equipped with VQGAN optimised with $2$ embedding methods on the LFW dataset): additional tables of results}

In the main part of the article, we presented the results of Experiment 2 without transfer for embedding method MagFace in Table \ref{lfw-5-fm-MagFace} and with transfer for embedding method SphereFace\ref{liu2017SphereFace} in Table \ref{vq_magface_mobilefacenet_sphereface}. Here we present another example without transfer for embedding method MobileFaceNet in Table \ref{lfw-5-fm-MobileFaceNet} and with transfer for embedding methods FaceNet, ArcFace, and ResNet\_152 in Tables \ref{vq_magface_mobilefacenet_facenet}, \ref{vq_magface_mobilefacenet_resnet} and \ref{vq_magface_mobilefacenet_arcface}.

From the results of the experiment $2$, we conclude that generative methods optimised with two different embeddings provide stronger privacy protection than those optimised with a single embedding method (as in experiment~$1$).

\subsection{Experiment 3 (comparing PrivacyGAN, AMT-GAN, and Fawkes on CC dataset): additional tables of results} \label{s2}
In this section, we present the results of experiment 3. All the tables here are similar to Table \ref{transfer_CC_sph} of the main paper, except for the differences in choice of transfer embeddings.

We present results for generative methods with a criterion based on transfer from MobileFaceNet and MagFace (used in our privacy algorithm) to embeddings FaceNet, ArcFace and ResNet\_152 (used in the recognition) in Tables \ref{transfer_CC_facenet}, \ref{transfer_CC_arcface} and \ref{transfer_CC_ResNet_152} as well as results without transfer for embeddings MagFace and MobileFaceNet in Tables \ref{transfer_CC_magface} and \ref{transfer_CC_facxmobile}.

Overall, results for the $CC$ dataset are similar to those for $LFW$, and generative methods remain preferable for privacy protection.

\subsection{Image examples for original and modified images using both pixel-based and generative methods}

The examples in this section show that generative methods modify image features more than the pixel-based methods. Nonetheless, they have less artificial pixel noise, which is common for images protected by Fawkes. Pixel noise can be more detrimental in terms of visual quality.

\subsubsection{Modified image examples: experiment 1}\label{s3}

\begin{figure*}[htbp] \begin{center}
\par
{\includegraphics[width=0.6\textwidth]{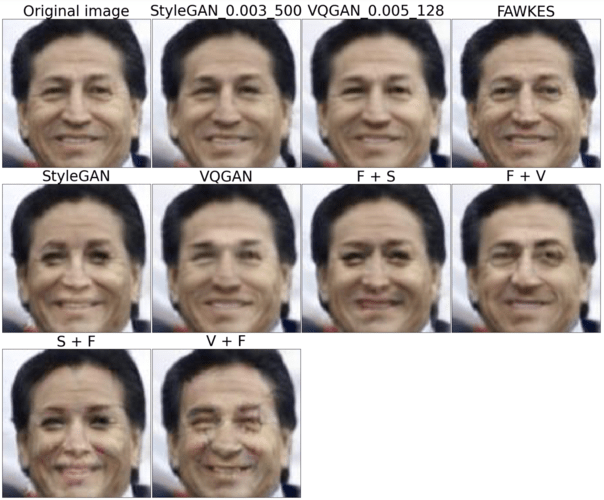}}%
\hfill
{\includegraphics[width=0.6\textwidth]{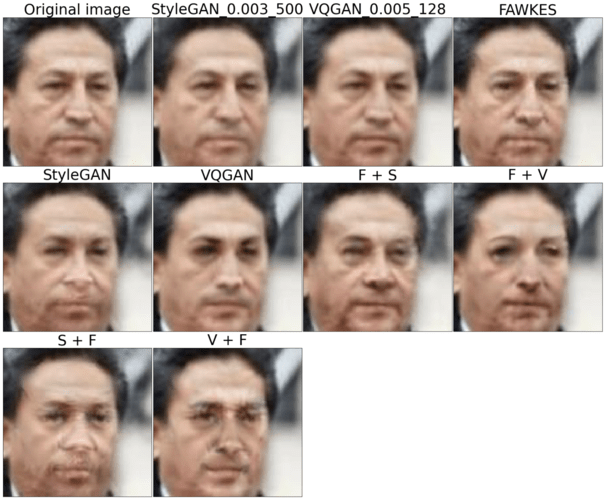} }%
\hfill
\par
\caption{\label{lfwf_1}
Experiment 1: Examples of original images from the LFW dataset and their counterparts modified by different privacy methods: StyleGAN\_0.003\_500, VQGAN\_0.005\_128, StyleGAN, VQGAN (using FaceNet as an embedding method for optimisation), Fawkes, and combinations of Fawkes with generative methods. Here we can see that while generative methods in general add more modification to an image than Fawkes, generative methods produce realistic images and do not add pixel noise.}
\end{center}
\end{figure*}

\begin{figure*}[htbp] \begin{center}
\par
{\includegraphics[width=0.6\textwidth]{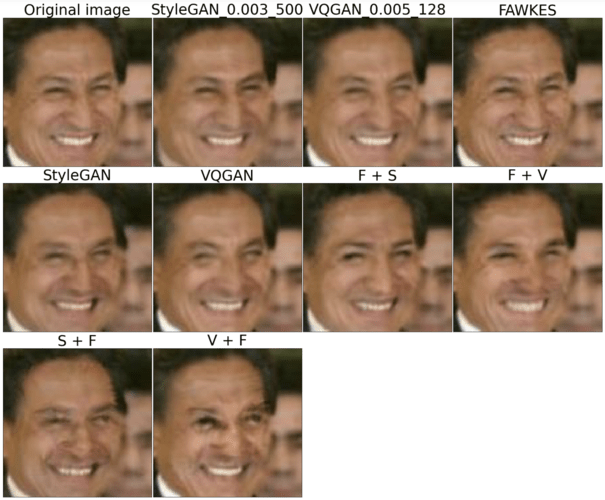} }%
\hfill
{\includegraphics[width=0.6\textwidth]{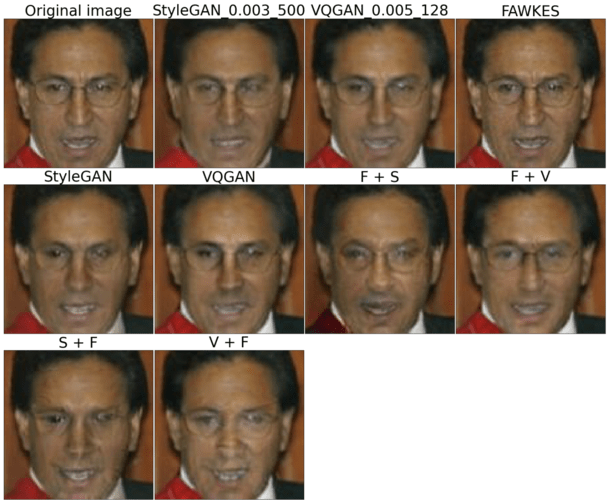} }%
\par
\caption{\label{lfwf_1_}
Experiment 1: Examples of original images from the LFW dataset and their counterparts modified by different privacy methods: StyleGAN\_0.003\_500, VQGAN\_0.005\_128, StyleGAN, VQGAN (using FaceNet as an embedding method for optimisation), Fawkes, and combinations of Fawkes with generative methods. Here we can see that while generative methods in general add more modification to an image than Fawkes, generative methods produce realistic images and do not add pixel noise.}
\end{center}
\end{figure*}

\begin{figure*}[htbp] \begin{center}
\par
{\includegraphics[width=0.6\textwidth]{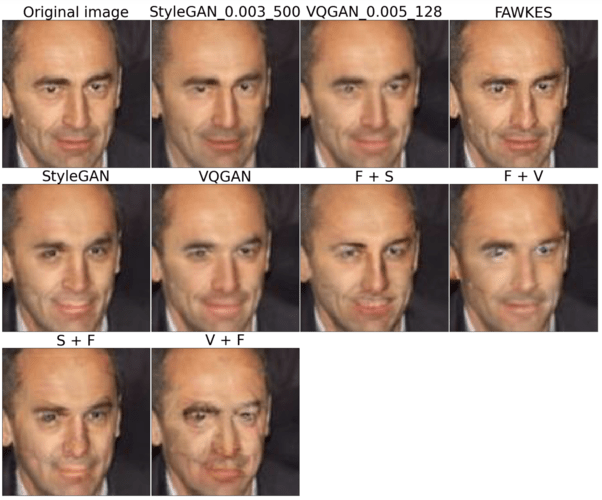}}%
\hfill
{\includegraphics[width=0.6\textwidth]{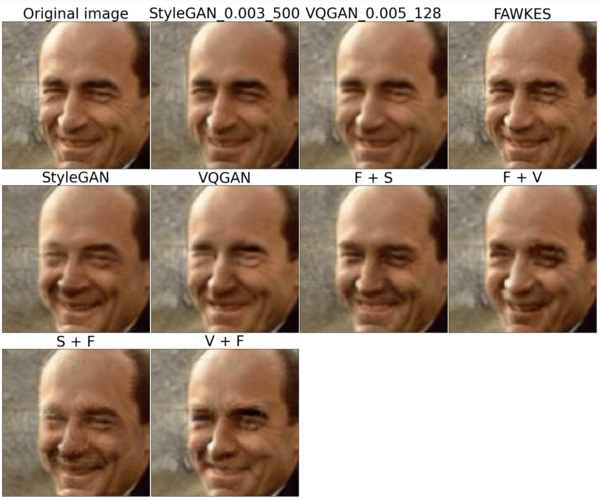} }%
\par
\caption{\label{lfwf_2}
Experiment 1: Examples of original images from the LFW dataset and their counterparts modified by different privacy methods: StyleGAN\_0.003\_500, VQGAN\_0.005\_128, StyleGAN, VQGAN (using FaceNet as an embedding method for optimisation), Fawkes, and combinations of Fawkes with generative methods. Here we can see that while generative methods in general add more modification to an image than Fawkes, generative methods produce realistic images and do not add pixel noise.}
\end{center}
\end{figure*}

\begin{figure*}[htbp] \begin{center}
\par
{\includegraphics[width=0.6\textwidth]{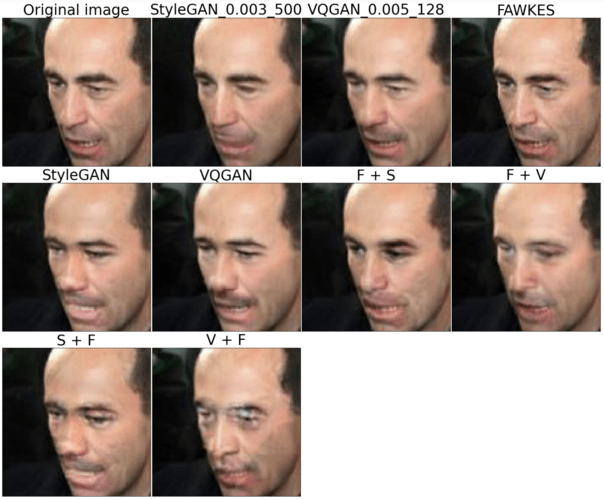} }%
\hfill
{\includegraphics[width=0.6\textwidth]{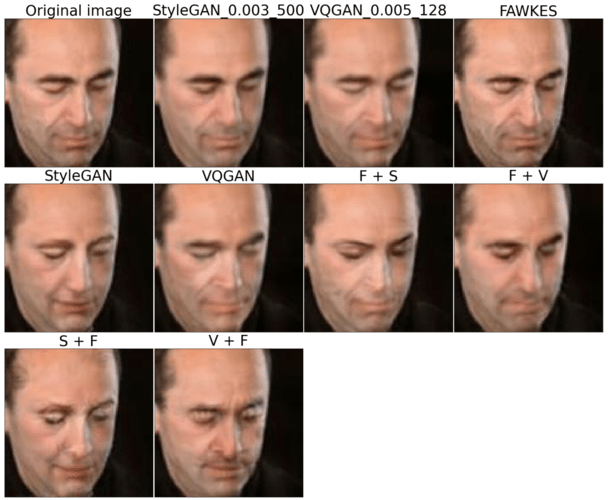} }%
\par
\caption{\label{lfwf_2_}
Experiment 1: Examples of original images from the LFW dataset and their counterparts modified by different privacy methods: StyleGAN\_0.003\_500, VQGAN\_0.005\_128, StyleGAN, VQGAN (using FaceNet as an embedding method for optimisation), Fawkes, and combinations of Fawkes with generative methods. Here we can see that while generative methods in general add more modification to an image than Fawkes, generative methods produce realistic images and do not add pixel noise.}
\end{center}
\end{figure*}

In this section, we present original images and their private versions that were obtained in the course of experiment 1 (similarly to {Fig.} \ref{image_facenet}). In addition, we also provide image examples for combinations of Fawkes and generative methods, namely Fawkes + StyleGAN (F+S), Fawkes + VQGAN (F+V), StyleGAN + Fawkes (S+F) and VQGAN + Fawkes (V+F). We see that adding Fawkes on top of generative methods improves image privacy, as in methods StyleGAN + Fawkes (S+F) and VQGAN + Fawkes (V+F). Tthe mentioned image examples can be found in Figs \ref{lfwf_1}, \ref{lfwf_1_}, \ref{lfwf_2} and \ref{lfwf_2_}.

\subsubsection{Modified image examples: experiment 2}

\begin{figure*}[htbp] \begin{center}
\par
{\includegraphics[width=0.501\textwidth]{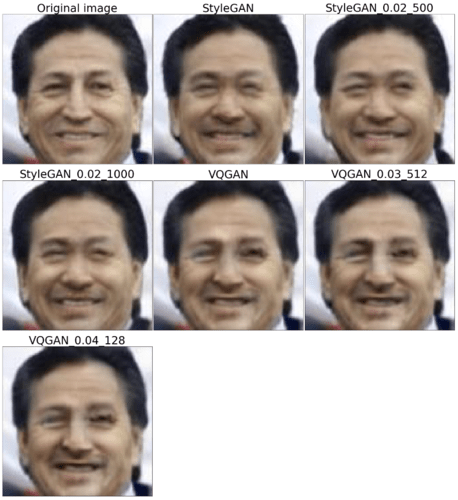}}%
\hfill
{\includegraphics[width=0.501\textwidth]{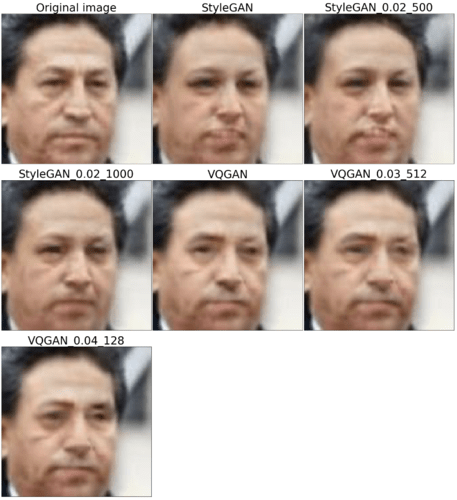} }%
\hfill
\caption{\label{lfwmf_1}
Experiment 2: Examples of images from the LFW dataset: the original image and the image modified by different privacy methods: StyleGAN, StyleGAN\_0.02\_500, StyleGAN\_0.02\_1000, VQGAN, VQGAN\_0.03\_512, VQGAN\_0.04\_128 with embedding methods MagFace and MobileFaceNet. Here, we can see that in some cases, increasing the number of iterations in optimisation and modifying the coefficient $K$ of the embedding method can affect the quality of an image while improving its privacy protection.}
\end{center}
\end{figure*}

\begin{figure*}[htbp] \begin{center}
\par
{\includegraphics[width=0.501\textwidth]{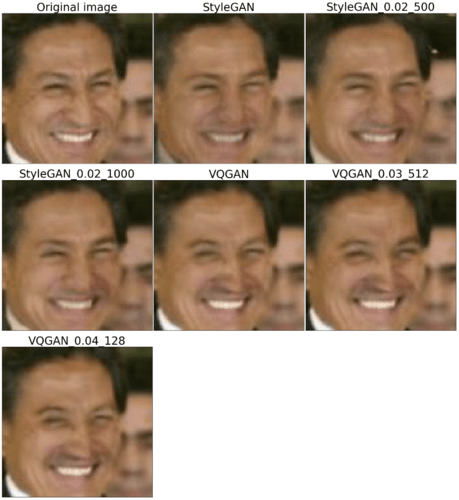} }%
\hfill
{\includegraphics[width=0.501\textwidth]{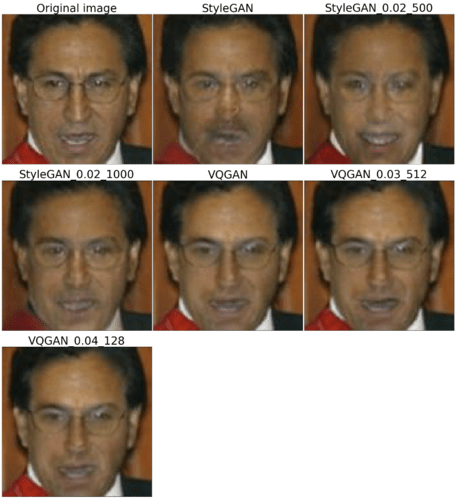} }%
\par
\caption{\label{lfwmf_1_}
Experiment 2: Examples of images from the LFW dataset: the original image and the image modified by different privacy methods: StyleGAN, StyleGAN\_0.02\_500, StyleGAN\_0.02\_1000, VQGAN, VQGAN\_0.03\_512, VQGAN\_0.04\_128 with embedding methods MagFace and MobileFaceNet. Here, we can see that in some cases, increasing the number of iterations in optimisation and modifying the coefficient $K$ of the embedding method can affect the quality of an image while improving its privacy protection.}
\end{center}
\end{figure*}

\begin{figure*}[htbp]
\begin{center}
\par
{\includegraphics[width=0.501\textwidth]{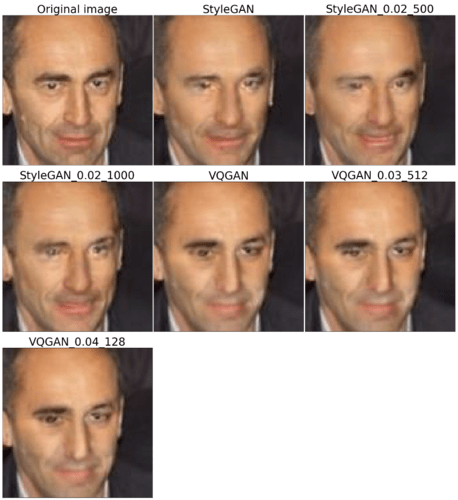}}%
\hfill
{\includegraphics[width=0.501\textwidth]{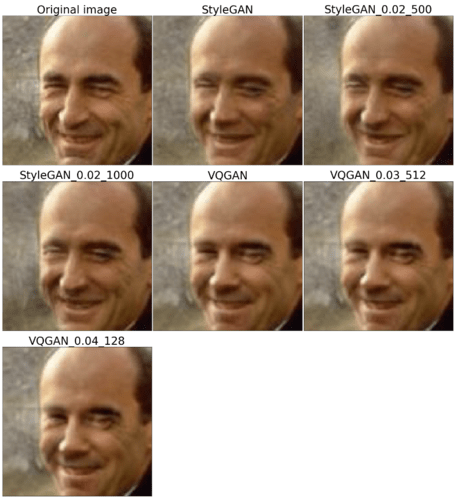} }%
\hfill
\par
\caption{\label{lfwmf_2}
Experiment 2: Examples of images from the LFW dataset: the original image and the image modified by different privacy methods: StyleGAN, StyleGAN\_0.02\_500, StyleGAN\_0.02\_1000, VQGAN, VQGAN\_0.03\_512, VQGAN\_0.04\_128 with embedding methods MagFace and MobileFaceNet. Here, we can see that in some cases, increasing the number of iterations in optimisation and modifying the coefficient $K$ of the embedding method can affect the quality of an image while improving its privacy protection.}
\end{center}
\end{figure*}

\begin{figure*}[htbp] \begin{center}
\par

{\includegraphics[width=0.501\textwidth]{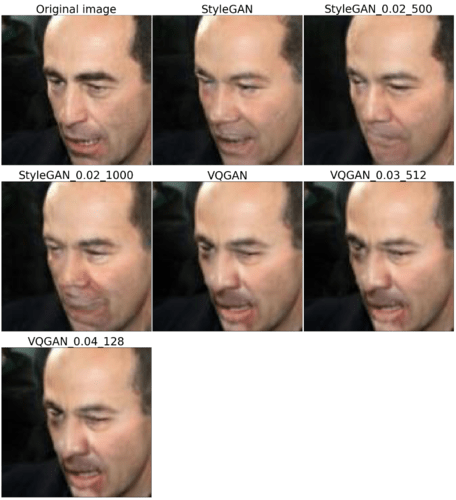} }%
\hfill
{\includegraphics[width=0.501\textwidth]{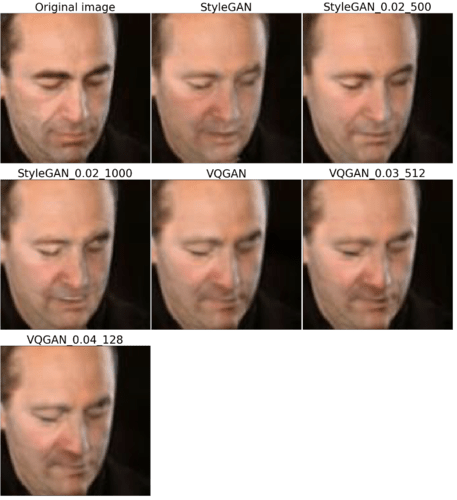} }%
\par
\caption{\label{lfwmf_2_}
Experiment 2: Examples of images from the LFW dataset: the original image and the image modified by different privacy methods: StyleGAN, StyleGAN\_0.02\_500, StyleGAN\_0.02\_1000, VQGAN, VQGAN\_0.03\_512, VQGAN\_0.04\_128 with embedding methods MagFace and MobileFaceNet. Here, we can see that in some cases, increasing the number of iterations in optimisation and modifying the coefficient $K$ of the embedding method can affect the quality of an image while improving its privacy protection.}
\end{center}
\end{figure*}

In this section, we present the original images and their private versions that were obtained during experiment 2, in addition to the images that were presented in the main paper in Fig \ref{image_magface_MobileFaceNet}.
These images can be found in Figs \ref{lfwmf_1}, \ref{lfwmf_1_}, \ref{lfwmf_2} and \ref{lfwmf_2_}.

\subsubsection{Modified image examples: experiment 3}\label{s4}

\begin{figure*}[htbp] \begin{center}
\par
{\includegraphics[width=0.501\textwidth]{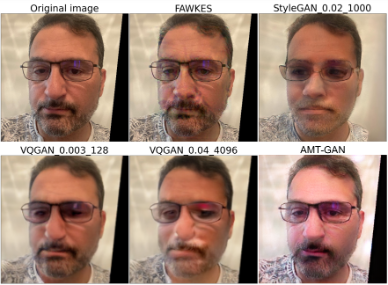} }
\hfill
{\includegraphics[width=0.501\textwidth]{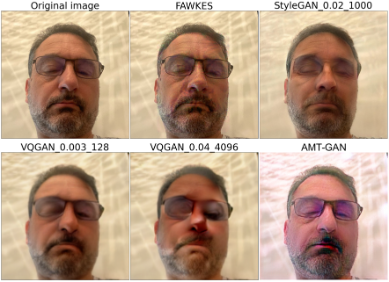} }%
\hfill
{\includegraphics[width=0.501\textwidth]{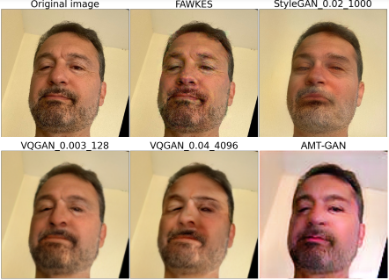} }%
\hfill
\par
\caption{\label{CC_supp_L}
Experiment 3: {Examples of images of volunteers modified by various privacy methods,} including AMT-GAN, Fawkes, StyleGAN\_0.02\_1000, VQGAN\_0.003\_128, VQGAN\_0.04\_4096 with embedding methods MagFace and MobileFaceNet. Here, we can see that in some cases, increasing the number of iterations in optimisation and modifying the coefficient $K$ of the embedding method can affect the quality of an image while improving its privacy protection. We can also see that some images are modified more than others after applying privacy-protection methods.}
\end{center}
\end{figure*}

\begin{figure*}[htbp] \begin{center}
\par
{\includegraphics[width=0.501\textwidth]{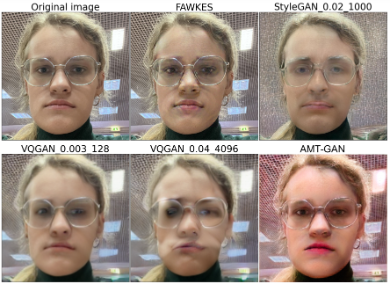} }%
\hfill
{\includegraphics[width=0.501\textwidth]{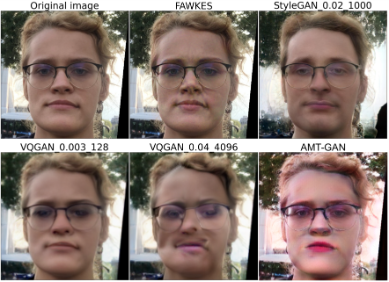} }%
\hfill
{\includegraphics[width=0.501\textwidth]{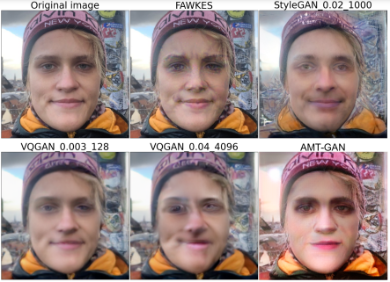} }%
\hfill
\par
\caption{\label{CC_supp_M}
Experiment 3: {Examples of images of volunteers modified by various privacy methods,} including AMT-GAN, Fawkes, StyleGAN\_0.02\_1000, VQGAN\_0.003\_128, VQGAN\_0.04\_4096 with embedding methods MagFace and MobileFaceNet. Here, we can see that in some cases, increasing the number of iterations in optimisation and modifying the coefficient $K$ of the embedding method can affect the quality of an image while improving its privacy protection. We can also see that some images are modified more than others after applying privacy-protection methods.}
\end{center}
\end{figure*}

\begin{figure*}[htbp]
\begin{center}
\par
{\includegraphics[width=0.501\textwidth]{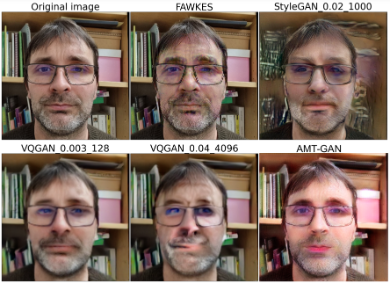} }%
\hfill
{\includegraphics[width=0.501\textwidth]{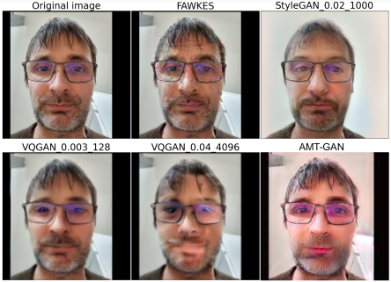} }%
\hfill
{\includegraphics[width=0.501\textwidth]{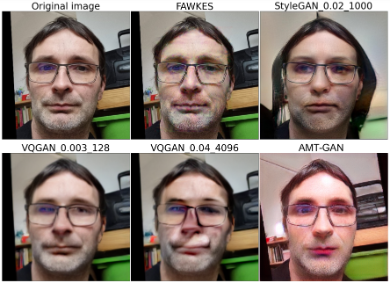} }%
\hfill
\par
\caption{\label{CC_supp_O}
Experiment 3: {Examples of images of volunteers modified by various privacy methods,} including AMT-GAN, Fawkes, StyleGAN\_0.02\_1000, VQGAN\_0.003\_128, VQGAN\_0.04\_4096 with embedding methods MagFace and MobileFaceNet. Here, we can see that in some cases, increasing the number of iterations in optimisation and modifying the coefficient $K$ of the embedding method can affect the quality of an image while improving its privacy protection. We can also see that some images are modified more than others after applying privacy-protection methods.}
\end{center}
\end{figure*}

Here, we present more image examples obtained by the procedure described in experiment 3. They are obtained the same way as images in Fig.~\ref{CC}. 
These examples are presented in Figs \ref{CC_supp_L}, \ref{CC_supp_M} and \ref{CC_supp_O}.

We would also like to note that all necessary approvals were obtained for the use of images in the present paper.
\begin{table*}[t]
	\begin{center}
	    \begin{adjustbox}{max width=\textwidth}
		
		\begin{tabular}{@{}lrrrr|r|rrrr@{}}
  \rowcolor{gray!30}
  \multicolumn{1}{c}{}&\multicolumn{4}{c}{PrivacyGAN}&\multicolumn{1}{|c|}{Pixel-based}& \multicolumn{4}{c}{Combinations}\\
			  & StyleGAN & StyleGAN & VQGAN & VQGAN & Fawkes & F + S & F + V & S + F & V + F \\
			  \rowcolor{gray!30}&&\_0.003\_500&&\_0.005\_128&&&&&\\
			\toprule
			Percentage  & 5.052 & 0.917 & 4.931 & 0.984 & 0.936 & 7.612 & 7.855 & \textbf{12.456} & 10.397 \\
 
			 \rowcolor{gray!30}Recall@1: m.i. & 7.962 & 23.981 & 8.531 & 22.464 & 23.602 & 5.782 & 5.118 & \textbf{2.464} & 3.791 \\
			Recall@1: o.i. & 8.246 & 24.976 & 10.095 & 24.692 & 24.313 & 6.445 & 7.062 & \textbf{2.749} & 4.408 \\
 
			 \rowcolor{gray!30}Recall@3: m.i. & 18.152 & 56.019 & 18.720 & 57.062 & 59.953 & 12.749 & 11.848 & \textbf{7.346} & 10.095 \\
			Recall@3: o.i. & 17.536 & 56.730 & 19.858 & 58.578 & 59.431 & 11.991 & 13.175 & \textbf{6.019} & 8.768 \\
 
			 \rowcolor{gray!30}Recall@5: m.i. & 24.076 & 69.905 & 24.265 & 72.986 & 76.209 & 16.919 & 16.114 & \textbf{10} & 13.081 \\
			Recall@5: o.i. & 22.038 & 70.379 & 24.929 & 72.559 & 75.877 & 15.782 & 16.445 & \textbf{8.199} & 11.611 \\
 
			 \rowcolor{gray!30}Recall@10: m.i. & 32.227 & 79.052 & 33.365 & 79.953 & 83.175 & 23.649 & 23.791 & \textbf{13.886} & 19.194 \\
			Recall@10: o.i. & 30.190 & 78.389 & 33.318 & 79.905 & 83.033 & 21.754 & 22.607 & \textbf{12.038} & 16.635 \\
 
			 \rowcolor{gray!30}Recall@50: m.i. & 53.791 & 90.379 & 55.829 & 90.995 & 92.464 & 42.559 & 43.223 & \textbf{28.436} & 34.882 \\
			Recall@50: o.i. & 52.701 & 91.185 & 55.924 & 91.469 & 92.227 & 41.706 & 42.607 & \textbf{26.398} & 31.848 \\
 
			 \rowcolor{gray!30}Recall@100: m.i. & 63.934 & 93.602 & 64.882 & 93.697 & 94.929 & 51.991 & 52.938 & \textbf{37.488} & 44.313 \\
			Recall@100: o.i. & 63.318 & 94.076 & 65.877 & 94.123 & 94.692 & 51.754 & 53.602 & \textbf{35.782} & 42.607 \\
						\bottomrule
		\end{tabular}
		\end{adjustbox}
	\end{center}
\caption{\label{st_vq_facenet_arcface} Evaluation on LFW dataset in the case of transfer to another embedding: VQGAN and StyleGAN are optimised with FaceNet, and all tests are performed with ArcFace. Lower recall and a higher percentage mean better privacy. As shown in Table \ref{vq_magface_mobilefacenet_arcface} methods that are optimised for two embeddings have a better transfer recall. We see that adding Fawkes on top of generative methods improves image privacy, as in the methods StyleGAN + Fawkes (S+F) and VQGAN + Fawkes (V+F).}
\end{table*}
\begin{table*}[t]
	\begin{center}
	    \begin{adjustbox}{max width=\textwidth}
		
		\begin{tabular}{@{}lrrrr|r|rrrr@{}}
  \rowcolor{gray!30}
    \multicolumn{1}{c}{}&\multicolumn{4}{c}{PrivacyGAN}&\multicolumn{1}{|c|}{Pixel-based}& \multicolumn{4}{c}{Combinations}\\
			  & StyleGAN & StyleGAN & VQGAN & VQGAN & Fawkes & F + S & F + V & S + F & V + F \\
			   \rowcolor{gray!30}& &\_0.003\_500& & \_0.005\_128&&&&&\\
			\toprule
			Percentage  & 1.226 & 0.446 & 0.947 & 0.454 & 0.454 & 1.875 & 4.359 & \textbf{8.645} & 5.586 \\
 
			 \rowcolor{gray!30}Recall@1: m.i. & 19.479 & 26.872 & 21.991 & 26.588 & 25.592 & 17.773 & \textbf{1.185} & 7.062 & 10.332 \\
			Recall@1: o.i. & 19.100 & 26.303 & 21.943 & 26.493 & 26.351 & 19.147 & 12.938 & \textbf{5.071} & 8.910 \\
 
			 \rowcolor{gray!30}Recall@3: m.i. & 55.166 & 72.275 & 58.815 & 73.981 & 72.986 & 43.033 & 18.436 & \textbf{16.825} & 23.555 \\
			Recall@3: o.i. & 51.090 & 74.692 & 56.682 & 72.701 & 73.697 & 41.517 & 26.161 & \textbf{11.232} & 19.005 \\
 
			 \rowcolor{gray!30}Recall@5: m.i. & 70.616 & 94.360 & 74.265 & 96.209 & 94.408 & 55.261 & 30.853 & \textbf{22.275} & 30.444 \\
			Recall@5: o.i. & 64.882 & 93.934 & 71.896 & 95.498 & 93.934 & 51.896 & 31.801 & \textbf{14.929} & 24.550 \\
 
			 \rowcolor{gray!30}Recall@10: m.i. & 77.678 & 96.303 & 82.227 & 97.536 & 96.351 & 65.118 & 42.227 & \textbf{29.052} & 38.341 \\
			Recall@10: o.i. & 75.024 & 96.066 & 80.379 & 97.156 & 96.019 & 61.848 & 41.185 & \textbf{21.801} & 33.033 \\
 
			 \rowcolor{gray!30}Recall@50: m.i. & 88.626 & 98.436 & 91.517 & 98.626 & 97.962 & 79.763 & 62.370 & \textbf{48.483} & 58.104 \\
			Recall@50: o.i. & 87.867 & 98.389 & 90.284 & 98.389 & 97.725 & 79.242 & 61.943 & \textbf{41.280} & 53.460 \\
 
			 \rowcolor{gray!30}Recall@100: m.i. & 91.659 & 98.815 & 94.313 & 98.815 & 98.436 & 84.929 & 71.090 & \textbf{55.735} & 66.446 \\
			Recall@100: o.i. & 91.754 & 98.673 & 93.223 & 98.768 & 98.152 & 84.408 & 69.716 & \textbf{51.043} & 62.844 \\
						\bottomrule
		\end{tabular}
		\end{adjustbox}
	\end{center}
\caption{\label{st_vq_facenet_mobilefacenet}Evaluation on LFW dataset in the case of transfer to another embedding: VQGAN and StyleGAN are optimised with FaceNet, and recognition (for all methods) is tested with MobileFaceNet. Lower recall and a higher percentage mean better privacy. Generative methods do obtain better results than Fawkes. We can also see that adding Fawkes on top of generative methods improves image privacy, as in the methods StyleGAN + Fawkes (S+F) and VQGAN + Fawkes (V+F). }
\end{table*}

\begin{table*}[t]
	\begin{center}
	    \begin{adjustbox}{max width=\textwidth}
		
		\begin{tabular}{@{}lrrrr|r|rrrr@{}}
  \rowcolor{gray!30}
      \multicolumn{1}{c}{}&\multicolumn{4}{c}{PrivacyGAN}&\multicolumn{1}{|c|}{Pixel-based}& \multicolumn{4}{c}{Combinations}\\
			  & StyleGAN & StyleGAN & VQGAN & VQGAN & Fawkes & F + S & F + V & S + F & V + F \\
			  \rowcolor{gray!30}&&\_0.003\_500&&\_0.005\_128&&&&&\\
			\toprule
			Percentage  & 0.670 & 0.342 & 0.564 & 0.395 & 0.408 & 1.273 & 2.573 & \textbf{6.823} & 3.309 \\
 
			 \rowcolor{gray!30}Recall@1: m.i. & 20.900 & 25.071 & 23.270 & 26.019 & 25.403 & 17.488 & 8.436 & \textbf{7.536} & 11.896 \\
			Recall@1: o.i. & 20.616 & 26.398 & 22.607 & 25.972 & 27.488 & 22.938 & 17.204 & \textbf{8.957} & 15.071 \\
 
			 \rowcolor{gray!30}Recall@3: m.i. & 59.526 & 73.128 & 65.592 & 75.545 & 76.351 & 49.573 & 30.000 & \textbf{18.957} & 34.597 \\
			Recall@3: o.i. & 58.199 & 73.886 & 63.886 & 72.559 & 75.498 & 49.431 & 38.009 & \textbf{17.251} & 31.706 \\
 
			 \rowcolor{gray!30}Recall@5: m.i. & 78.673 & 97.773 & 87.583 & 98.626 & 98.578 & 66.303 & 45.498 & \textbf{27.299} & 45.924 \\
			Recall@5: o.i. & 73.791 & 97.441 & 83.081 & 98.531 & 98.389 & 62.512 & 48.720 & \textbf{21.706} & 40.758 \\
 
			 \rowcolor{gray!30}Recall@10: m.i. & 85.972 & 98.483 & 91.801 & 98.863 & 99.005 & 74.976 & 58.483 & \textbf{35.545} & 56.493 \\
			Recall@10: o.i. & 82.891 & 98.389 & 89.668 & 98.863 & 98.815 & 72.370 & 59.479 & \textbf{30.332} & 51.754 \\
 
			 \rowcolor{gray!30}Recall@50: m.i. & 94.028 & 99.005 & 97.014 & 99.052 & 99.194 & 87.915 & 78.152 & \textbf{55.735} & 74.123 \\
			Recall@50: o.i. & 93.128 & 99.147 & 95.924 & 99.194 & 99.147 & 87.583 & 77.393 & \textbf{51.469} & 72.464 \\
 
			 \rowcolor{gray!30}Recall@100: m.i. & 95.877 & 99.100 & 97.867 & 99.100 & 99.194 & 91.659 & 83.744 & \textbf{63.934} & 80.095 \\
			Recall@100: o.i. & 95.308 & 99.194 & 97.299 & 99.194 & 99.147 & 91.422 & 82.938 & \textbf{60.900} & 78.531 \\
						\bottomrule
		\end{tabular}
		\end{adjustbox}
	\end{center}
\caption{ \label{vq_st_facenet_resnet}Evaluation on LFW dataset in the case of transfer to another embedding: VQGAN and StyleGAN are optimised with FaceNet and tested with ResNet\_152. Lower recall and a higher percentage mean better privacy. The generative methods with two embeddings (see Table \ref{vq_magface_mobilefacenet_resnet}) do obtain better results, showing that using multiple embeddings increases robustness and transfer. We can also see that adding Fawkes on top of generative methods improves image privacy, as in the methods StyleGAN + Fawkes (S+F) and VQGAN + Fawkes (V+F). }
\end{table*}

\begin{table*}[t]
	\begin{center}
	\begin{adjustbox}{max width=\textwidth}

		\begin{tabular}{@{}lrrrrrrrrr@{}}
  \rowcolor{gray!30}
			  & StyleGAN & StyleGAN & StyleGAN  & VQGAN & VQGAN  & VQGAN  \\
			   \rowcolor{gray!30}			  &   &   \_0.02\_500 &  \_0.02\_1000 &   &  \_0.03\_512 &  \_0.04\_128 \\
			\hline

			\hline
			Percentage  & \textbf{7.849} & 4.494 & 4.107 & 2.626 & 3.470 & 3.807 \\
 
			\rowcolor{gray!30}			  Recall@1: m.i. & \textbf{2.844} & 6.066 & 6.730 & 8.152 & 6.919 & 6.019 \\
			Recall@1: o.i. & \textbf{4.597} & 8.294 & 10.047 & 11.706 & 9.479 & 9.431 \\
 
			\rowcolor{gray!30}			  Recall@3: m.i. & \textbf{9.242} & 15.640 & 18.009 & 22.701 & 19.242 & 17.536 \\
			Recall@3: o.i. & \textbf{9.668} & 17.156 & 19.479 & 25.545 & 20.995 & 19.431 \\
 
			\rowcolor{gray!30}			  Recall@5: m.i. & \textbf{13.175} & 22.322 & 25.166 & 32.275 & 27.062 & 24.787 \\
			Recall@5: o.i. & \textbf{13.081} & 22.701 & 25.118 & 32.749 & 27.109 & 25.592 \\
 
			\rowcolor{gray!30}			  Recall@10: m.i. & \textbf{18.578} & 31.564 & 35.118 & 43.175 & 37.725 & 34.408 \\
			Recall@10: o.i. & \textbf{17.725} & 30.758 & 34.360 & 43.412 & 36.019 & 33.460 \\
 
			\rowcolor{gray!30}			  Recall@50: m.i. & \textbf{37.441} & 54.028 & 57.062 & 67.725 & 60.142 & 57.441 \\
			Recall@50: o.i. & \textbf{35.213} & 52.464 & 55.118 & 66.019 & 58.910 & 55.592 \\
 
			\rowcolor{gray!30}			  Recall@100: m.i. & \textbf{48.531} & 64.882 & 68.436 & 77.109 & 71.137 & 68.104 \\
			Recall@100: o.i. & \textbf{46.351} & 63.934 & 65.735 & 75.640 & 69.147 & 67.014 \\
			\bottomrule
		\end{tabular}
	    \end{adjustbox}

	\end{center}
\caption{ \label{vq_magface_mobilefacenet_facenet} Evaluation on LFW dataset in the case of transfer to another embedding: VQGAN and StyleGAN are optimised with MagFace and MobileFaceNet, and recognition is tested with FaceNet. Lower recall and a higher percentage mean better privacy. In this case, generative methods optimised with two embeddings obtain worse results than generative methods optimised with only one embedding method in Table \ref{facenet_facenet}: this is, however, not a fair comparison because we do not use FaceNet for optimisation in this experiment, while we do in Table \ref{facenet_facenet}.}
\end{table*}

\begin{table*}[t]
	\begin{center}
	    \begin{adjustbox}{max width=\textwidth}

		\begin{tabular}{@{}lrrrrrrrrr@{}}
  \rowcolor{gray!30}

			  & StyleGAN & StyleGAN & StyleGAN  & VQGAN & VQGAN  & VQGAN  \\
			   \rowcolor{gray!30}			  &   &   \_0.02\_500 &  \_0.02\_1000 &   &  \_0.03\_512 &  \_0.04\_128 \\
			\toprule
			Percentage:  & \textbf{5.656} & 2.671 & 2.296 & 1.549 & 2.097 & 2.605 \\
 
			  \rowcolor{gray!30}Recall@1: m.i. & \textbf{4.976} & 8.294 & 11.422 & 12.322 & 9.526 & 7.630 \\
			Recall@1: o.i. & \textbf{9.289} & 15.450 & 18.152 & 18.957 & 16.635 & 15.403 \\
 
			  \rowcolor{gray!30}Recall@3: m.i. & \textbf{15.829} & 28.578 & 32.986 & 37.488 & 31.043 & 28.057 \\
			Recall@3: o.i. & \textbf{19.763} & 34.597 & 39.100 & 44.455 & 38.863 & 35.545 \\
 
			  \rowcolor{gray!30}Recall@5: m.i. & \textbf{23.507} & 41.327 & 46.445 & 55.687 & 46.161 & 41.469 \\
			Recall@5: o.i. & \textbf{26.114} & 43.175 & 49.858 & 56.303 & 48.246 & 43.981 \\
 
			  \rowcolor{gray!30}Recall@10: m.i. & \textbf{34.218} & 52.796 & 58.957 & 68.673 & 59.716 & 54.929 \\
			Recall@10: o.i. & \textbf{34.313} & 52.796 & 60.616 & 68.531 & 60.284 & 55.071 \\
 
			  \rowcolor{gray!30}Recall@50: m.i. & \textbf{56.398} & 74.123 & 78.104 & 84.692 & 79.763 & 76.540 \\
			Recall@50: o.i. & \textbf{54.929} & 73.507 & 79.005 & 84.218 & 79.242 & 77.109 \\
 
			  \rowcolor{gray!30}Recall@100: m.i. & \textbf{64.882} & 81.611 & 84.360 & 89.242 & 85.118 & 82.986 \\
			Recall@100: o.i. & \textbf{63.839} & 80.853 & 84.550 & 88.863 & 85.118 & 83.081 \\
			\bottomrule
		\end{tabular}
    \end{adjustbox}
	\end{center}
\caption{Evaluation on LFW dataset in the case of transfer to another embedding: VQGAN and StyleGAN are optimised with MagFace and MobileFaceNet and recognition is tested with ResNet\_152\label{vq_magface_mobilefacenet_resnet}. Lower recall and a higher percentage mean better privacy. In this case, generative methods optimised with two embeddings obtain better results than generative methods optimised with only one embedding method, as in Table \ref{st_vq_facenet_mobilefacenet}. }
\end{table*}

\begin{table*}[t]
	\begin{center}
	    \begin{adjustbox}{max width=\textwidth}

		\begin{tabular}{@{}lrrrrrrrrr@{}}
  \rowcolor{gray!30}
			  & StyleGAN & StyleGAN & StyleGAN & VQGAN & VQGAN & VQGAN \\
			 \rowcolor{gray!30} &&\_0.02\_500&\_0.02\_1000&&\_0.03\_512&\_0.04\_128\\
			\toprule
			Percentage  & \textbf{11.993} & 8.305 & 7.534 & 6.067 & 7.164 & 8.181 \\
 
			  \rowcolor{gray!30}Recall@1: m.i. & \textbf{2.038} & 4.123 & 4.550 & 6.967 & 4.882 & 4.739 \\
			Recall@1: o.i. & \textbf{3.555} & 5.687 & 7.488 & 7.867 & 6.540 & 6.351 \\
 
			  \rowcolor{gray!30}Recall@3: m.i. & \textbf{6.588} & 11.754 & 12.370 & 15.640 & 13.270 & 11.232 \\
			Recall@3: o.i. & \textbf{6.919} & 12.417 & 14.787 & 16.398 & 13.791 & 12.038 \\
 
			  \rowcolor{gray!30}Recall@5: m.i. & \textbf{9.005} & 16.209 & 17.109 & 20.758 & 19.052 & 16.588 \\
			Recall@5: o.i. & \textbf{9.526} & 16.493 & 18.673 & 21.611 & 18.578 & 16.066 \\
 
			  \rowcolor{gray!30}Recall@10: m.i. & \textbf{13.081} & 22.844 & 24.976 & 29.384 & 26.066 & 22.417 \\
			Recall@10: o.i. & \textbf{14.171} & 22.938 & 25.450 & 29.810 & 25.450 & 22.322 \\
 
			  \rowcolor{gray!30}Recall@50: m.i. & \textbf{27.299} & 41.374 & 43.270 & 51.280 & 44.455 & 41.611 \\
			Recall@50: o.i. & \textbf{28.863} & 41.991 & 45.024 & 51.754 & 46.730 & 42.464 \\
 
			  \rowcolor{gray!30}Recall@100: m.i. & \textbf{36.445} & 50.806 & 52.986 & 61.137 & 54.360 & 51.232 \\
			Recall@100: o.i. & \textbf{38.910} & 53.128 & 56.161 & 62.796 & 57.867 & 52.796 \\
			\bottomrule
		\end{tabular}
    \end{adjustbox}
	\end{center}
\caption{Evaluation on LFW dataset, in the case of transfer to another embedding: VQGAN and StyleGAN are optimised with MagFace and MobileFaceNet, and recognition is tested with ArcFace. Lower recall and a higher percentage mean better privacy. \label{vq_magface_mobilefacenet_arcface}  In this case, generative methods optimised with two embeddings obtain better results than generative methods optimised with only one embedding method, as in Table \ref{st_vq_facenet_arcface}.
}
\end{table*}

\begin{table*}[htbp]
	\begin{center}
	    \begin{adjustbox}{max width=\textwidth}

		\begin{tabular}{@{}lrrrrrrrrr@{}}
  \rowcolor{gray!30}
			  & StyleGAN & StyleGAN & StyleGAN & VQGAN & VQGAN & VQGAN \\
			 \rowcolor{gray!30} &&\_0.02\_500&\_0.02\_1000&&\_0.03\_512&\_0.04\_128\\
			\toprule
			Percentage  & \textbf{6.925} & 3.290 & 3.079 & 2.583 & 3.555 & 4.461 \\
 
			  \rowcolor{gray!30}Recall@1: m.i. & \textbf{0.569} & 2.986 & 2.749 & 2.227 & 0.900 & 0.664 \\
			Recall@1: o.i. & \textbf{6.635} & 14.739 & 14.976 & 15.782 & 12.654 & 10.379 \\
 
			  \rowcolor{gray!30}Recall@3: m.i. & \textbf{11.090} & 22.417 & 24.408 & 26.066 & 20.190 & 17.062 \\
			Recall@3: o.i. & \textbf{13.365} & 30.095 & 32.180 & 34.834 & 26.351 & 22.085 \\
 
			  \rowcolor{gray!30}Recall@5: m.i. & \textbf{18.531} & 36.635 & 38.152 & 42.607 & 31.517 & 28.246 \\
			Recall@5: o.i. & \textbf{18.009} & 38.578 & 40.379 & 45.308 & 34.550 & 29.289 \\
 
			  \rowcolor{gray!30}Recall@10: m.i. & \textbf{26.682} & 48.389 & 49.953 & 55.735 & 43.223 & 39.052 \\
			Recall@10: o.i. & \textbf{24.739} & 48.720 & 49.100 & 54.739 & 44.645 & 38.720 \\
 
			  \rowcolor{gray!30}Recall@50: m.i. & \textbf{47.014} & 69.384 & 71.611 & 74.171 & 65.687 & 59.668 \\
			Recall@50: o.i. & \textbf{46.872} & 69.716 & 71.991 & 74.597 & 66.919 & 60.427 \\
 
			  \rowcolor{gray!30}Recall@100: m.i. & \textbf{56.682} & 77.583 & 78.957 & 80.237 & 73.981 & 68.578 \\
			Recall@100: o.i. & \textbf{56.967} & 76.588 & 79.336 & 80.948 & 75.308 & 69.431 \\
   \bottomrule
		\end{tabular}
     \end{adjustbox}
	\end{center}
\caption{Evaluation on LFW dataset: VQGAN and StyleGAN (and their variants) are optimised with MagFace and MobileFaceNet, and recognition is tested with MobileFaceNet. Generative methods do obtain better privacy than Fawkes (Table \ref{st_vq_facenet_mobilefacenet}). \label{lfw-5-fm-MobileFaceNet}
}
\end{table*}

\begin{table*}[htbp]
	\begin{center}
	\begin{adjustbox}{max width=\textwidth}

		\begin{tabular}{@{}ccHHHHcHHHHHHHHHccHHcHc@{}}
  \rowcolor{gray!30}
         & VQGAN& VQGAN & VQGAN\_0.007\_128 & VQGAN & VQGAN& VQGAN & VQGAN & VQGAN & VQGAN & VQGAN & VQGAN & VQGAN & VQGAN & VQGAN & VQGAN& VQGAN & Fawkes & Fawkes & Fawkes & StyleGAN & StyleGAN & AMT-GAN\\
	 \rowcolor{gray!30}		  			&  \_0.003\_128 & \_0.005\_128 & \_0.007\_128 & \_0.01\_128 & \_0.02\_128 && \_0.03\_256 & \_0.03\_512 & \_0.03\_1024 & \_0.03\_2048 & \_0.04\_128 & \_0.04\_256 & \_0.04\_512 & \_0.04\_1024 & \_0.04\_2048 & \_0.04\_4096 &  & \_{high4} & \_{high5} & \_0.02\_1000 & \_0.01\_500 & \\
			\hline
			Percentage & 0.697 & 0.743 & 0.839 & 1.102 & 2.395 & 3.975 & 4.466 & 5.148 & 5.599 & 5.791 & 5.438 & 6.371 & 7.035 & 7.477 & 8.034 & 8.312 & 0.715 & 16.483 & 15.159 & \textbf{12.268} & \textbf{19.590} & 1.633 \\
			  \rowcolor{gray!30}Recall@1: m.i. & 23.571 & 22.869 & 21.886 & 18.756 & 13.721 & 10.712 & 9.107 & 8.566 & 7.462 & 7.302 & 8.004 & 6.820 & 5.978 & 5.316 & 5.256 & 5.055 & 23.831 & 5.697 & 6.580 & \textbf{3.450} &\textbf{ 0.602} & 15.868\\
			Recall@1: o.i. & 24.012 & 23.992 & 23.370 & 22.648 & 18.415 & 15.125 & 14.002 & 11.856 & 10.973 & 10.792 & 11.434 & 10.171 & 9.288 & 8.526 & 7.362 & 8.064 & 24.534 & 6.700 & 6.881 &\textbf{ 6.239 }& \textbf{1.545} & 16.429\\
			  \rowcolor{gray!30}Recall@3: m.i. & 70.271 & 66.740 & 61.645 & 55.266 & 37.452 & 26.359 & 23.511 & 21.966 & 19.358 & 18.375 & 21.444 & 17.553 & 16.189 & 13.882 & 13.902 & 12.778 & 66.680 & 14.483 & 16.189 &\textbf{ 8.947} & \textbf{1.805 } &  45.256 \\
			Recall@3: o.i. & 71.274 & 68.265 & 62.909 & 57.593 & 40.983 & 32.177 & 28.365 & 25.797 & 23.571 & 23.049 & 25.135 & 22.287 & 19.498 & 17.793 & 16.329 & 15.527 & 66.800 & 14.524 & 15.627 & \textbf{12.237} & \textbf{3.671} &  46.319\\
			  \rowcolor{gray!30}Recall@5: m.i. & 91.775 & 86.841 & 80.662 & 72.317 & 49.529 & 36.169 & 32.638 & 30.090 & 26.881 & 25.998 & 29.408 & 24.433 & 22.708 & 19.759 & 19.157 & 17.934 & 87.964 & 19.338 & 21.344 & \textbf{13.340} & \textbf{2.969 } & 60.522\\
			Recall@5: o.i. & 91.555 & 86.861 & 80.301 & 71.374 & 51.294 & 39.860 & 35.547 & 32.859 & 30.271 & 28.947 & 31.856 & 28.405 & 24.694 & 23.330 & 20.802 & 20.040 & 84.875 & 18.395 & 19.679 & \textbf{15.787} &\textbf{ 4.774} & 60.923 \\
			  \rowcolor{gray!30}Recall@10: m.i. & 94.965 & 91.655 & 87.282 & 80.020 & 59.358 & 45.537 & 41.404 & 38.656 & 35.547 & 34.403 & 37.312 & 33.079 & 29.609 & 26.861 & 25.637 & 24.835 & 91.575 & 24.273 & 26.339 & \textbf{18.495} & \textbf{4.493} & 68.867\\
			Recall@10: o.i. & 94.905 & 92.257 & 87.161 & 79.519 & 60.502 & 48.265 & 44.594 & 40.843 & 37.693 & 37.011 & 40.281 & 36.088 & 32.818 & 29.930 & 27.964 & 26.239 & 89.950 & 23.009 & 24.072 & \textbf{20.963} & \textbf{6.841} & 69.829 \\
			  \rowcolor{gray!30}Recall@50: m.i. & 97.733 & 96.670 & 94.604 & 91.153 & 76.489 & 65.597 & 61.705 & 58.215 & 55.527 & 54.804 & 57.131 & 50.351 & 49.388 & 46.700 & 45.356 & 42.508 & 96.309 & 38.114 & 40.281 & \textbf{34.483} &\textbf{ 12.558 } & 83.470\\
			Recall@50: o.i. & 97.553 & 97.172 & 95.085 & 90.732 & 78.054 & 66.941 & 63.430 & 60.883 & 56.891 & 56.891 & 60.060 & 54.584 & 52.979 & 49.027 & 47.663 & 45.436 & 95.125 & 34.544 & 35.988 & \textbf{36.349} & \textbf{15.747 } & 84.554 \\
			  \rowcolor{gray!30}Recall@100: m.i. & 98.235 & 97.673 & 96.249 & 93.982 & 83.370 & 72.457 & 69.127 & 66.520 & 64.293 & 63.270 & 65.878 & 61.063 & 58.576 & 55.667 & 54.022 & 51.775 & 97.151 & 44.534 & 47.563 & \textbf{42.989} &\textbf{ 19.077 } & 88.465\\
			Recall@100: o.i. & 98.195 & 97.954 & 97.372 & 94.122 & 83.751 & 75.125 & 71.133 & 68.666 & 66.941 & 65.697 & 68.706 & 63.410 & 62.106 & 58.556 & 57.212 & 55.486 & 96.670 & 41.324 & 43.771 &\textbf{ 46.098} & \textbf{22.006 } & 89.749\\
		\bottomrule \end{tabular}
  \end{adjustbox}
	\end{center}
\caption{\label{transfer_CC_facenet} Evaluation on CC dataset in the case of a transfer to another embedding: VQGAN and StyleGAN are optimised with MagFace and MobileFaceNet, and recognition is tested with FaceNet. Lower recall and a higher percentage mean better privacy. We can see that generative methods' evaluation results for the CC dataset are very similar to the ones for the LFW dataset.}
\end{table*}

\begin{table*}[htbp]
	\begin{center}
	\begin{adjustbox}{max width=\textwidth}
		
		\begin{tabular}{@{}ccHHHHcHHHHHHHHHccHHcHc@{}}
  \rowcolor{gray!30}
			  & VQGAN& VQGAN & VQGAN\_0.007\_128 & VQGAN & VQGAN& VQGAN & VQGAN & VQGAN & VQGAN & VQGAN & VQGAN & VQGAN & VQGAN & VQGAN & VQGAN& VQGAN & Fawkes & Fawkes & Fawkes & StyleGAN & StyleGAN & AMT-GAN\\
			 \rowcolor{gray!30}  			&  \_0.003\_128 & \_0.005\_128 & \_0.007\_128 & \_0.01\_128 & \_0.02\_128 & & \_0.03\_256 & \_0.03\_512 & \_0.03\_1024 & \_0.03\_2048 & \_0.04\_128 & \_0.04\_256 & \_0.04\_512 & \_0.04\_1024 & \_0.04\_2048 & \_0.04\_4096 &  & \_{high4} & \_{high5} & \_0.02\_1000 & \_0.01\_500 & \\
			\hline
			Percentage & 2.880 & 3.137 & 3.537 & 4.669 & 9.723 & 14.337 & 16.945 & 18.769 & 20.616 & 21.291 & 18.475 & 21.202 & 23.513 & 25.043 & 26.442 & \textbf{27.089} & 2.507 & 7.877 & 7.835 & 22.651 & 6.891 & 4.201 \\
			  \rowcolor{gray!30}Recall@1: m.i. & 23.390 & 20.923 & 17.492 & 13.099 & 4.534 & 2.086 & 1.886 & 1.264 & 0.943 & 0.863 & 1.404 & 0.923 & 0.682 & 0.481 & {0.401} & 0.562 & 24.674 & 16.610 & 16.429 & \textbf{0.301 }& 10.291 & 19.097\\
			Recall@1: o.i. & 21.926 & 20.401 & 19.498 & 17.492 & 9.408 & 5.918 & 4.293 & 3.230 & 3.170 & 2.808 & 3.591 & 2.648 & 1.926 & 1.625 & 1.123 & \textbf{1.043} & 24.995 & 18.235 & 16.891 & 1.886 & 13.741 & 17.031 \\
			  \rowcolor{gray!30}Recall@3: m.i. & 62.708 & 55.246 & 47.001 & 34.564 & 13.139 & 5.998 & 4.754 & 3.571 & 3.069 & 2.889 & 4.333 & 2.608 & 1.886 & 1.204 & 1.244 & \textbf{ 1.143 }& 67.462 & 38.455 & 37.833 & 1.484 & 27.844 & 52.277\\
			Recall@3: o.i. & 63.129 & 57.573 & 50.612 & 41.404 & 19.458 & 10.973 & 8.285 & 6.179 & 5.797 & 5.537 & 6.921 & 5.095 & 3.872 & 2.969 & 2.548 & \textbf{2.187} & 67.683 & 39.278 & 38.275 & 3.992 & 30.732 & 49.107\\
			  \rowcolor{gray!30}Recall@5: m.i. & 78.335 & 70.431 & 60.722 & 46.740 & 19.418 & 9.468 & 7.583 & 5.797 & 5.055 & 4.313 & 6.199 & 4.072 & 3.009 & 2.227 & 2.046 &\textbf{ 1.805 }& 84.393 & 48.024 & 46.800 & 2.768 & 37.432 & 64.654\\
			Recall@5: o.i. & 78.154 & 71.936 & 62.387 & 50.532 & 24.594 & 13.561 & 10.311 & 7.663 & 7.081 & 6.961 & 8.606 & 6.479 & 5.296 & 3.992 & 3.290 &\textbf{ 2.949} & 83.952 & 46.520 & 45.898 & 5.296 & 38.716 & 63.149\\
			  \rowcolor{gray!30}Recall@10: m.i. & 81.846 & 76.269 & 68.365 & 56.891 & 27.623 & 14.845 & 11.655 & 9.569 & 8.325 & 7.422 & 10.070 & 6.941 & 5.376 & 4.253 & 3.611 & \textbf{3.230} & 86.399 & 53.721 & 52.197 & 4.975 & 45.416 & 70.351\\
			Recall@10: o.i. & 82.046 & 76.710 & 69.047 & 57.733 & 30.532 & 17.553 & 14.062 & 10.191 & 9.850 & 9.468 & 11.575 & 8.626 & 7.202 & 5.657 &\textbf{ 4.794} & \textbf{4.794 }& 86.219 & 52.718 & 51.856 & 7.763 & 46.419  & 68.706\\
			  \rowcolor{gray!30}Recall@50: m.i. & 87.182 & 84.694 & 79.940 & 72.237 & 45.155 & 30.211 & 23.731 & 20.201 & 18.175 & 17.192 & 20.441 & 16.510 & 13.240 & 11.294 & 10.672 & \textbf{9.007 }& 89.709 & 64.554 & 64.112 & 14.624 & 61.906 & 80.100\\
			Recall@50: o.i. & 87.603 & 84.413 & 80.060 & 71.956 & 47.121 & 31.033 & 25.537 & 21.324 & 18.937 & 17.874 & 22.267 & 17.773 & 15.005 & 12.798 & 11.755 & \textbf{11.013} & 89.749 & 64.493 & 63.390 & 16.309 & 62.969 & 78.495\\
			  \rowcolor{gray!30}Recall@100: m.i. & 89.087 & 87.503 & 84.032 & 77.332 & 53.661 & 38.736 & 31.595 & 27.282 & 24.855 & 23.611 & 28.164 & 22.708 & 19.699 & 16.931 & 15.928 & \textbf{14.203} & 90.853 & 70.050 & 69.288 & 20.702 & 69.188 & 84.092\\
			Recall@100: o.i. & 89.348 & 87.482 & 84.293 & 77.693 & 54.564 & 39.238 & 32.859 & 28.465 & 25.717 & 23.932 & 29.549 & 24.614 & 20.120 & 17.472 & 17.111 & \textbf{15.226} & 90.913 & 69.890 & 68.485 & 22.287 & 69.388 & 82.247\\
		\bottomrule \end{tabular} \end{adjustbox}
	\end{center}
\caption{\label{transfer_CC_magface} Evaluation on CC dataset, in the case without transfer to another embedding: VQGAN and StyleGAN are optimised with MagFace and MobileFaceNet, and recognition is tested with MagFace. We can see that generative methods' evaluation results for the CC dataset are very similar to the ones for the LFW dataset.}
\end{table*}

\begin{table*}[htbp]
	\begin{center}
	\begin{adjustbox}{max width=\textwidth}
		
		\begin{tabular}{@{}ccHHHHcHHHHHHHHHccHHcHc@{}}
  \rowcolor{gray!30}
			& VQGAN& VQGAN & VQGAN\_0.007\_128 & VQGAN & VQGAN& VQGAN & VQGAN & VQGAN & VQGAN & VQGAN & VQGAN & VQGAN & VQGAN & VQGAN & VQGAN& VQGAN & Fawkes & Fawkes & Fawkes & StyleGAN & StyleGAN & AMT-GAN \\
			 \rowcolor{gray!30}  			&  \_0.003\_128 & \_0.005\_128 & \_0.007\_128 & \_0.01\_128 & \_0.02\_128 & & \_0.03\_256 & \_.03\_512 & \_0.03\_1024 & \_0.03\_2048 & \_0.04\_128 & \_0.04\_256 & \_0.04\_512 & \_0.04\_1024 & \_0.04\_2048 & \_0.04\_4096 &  & \_{high4} & \_{high5} & \_0.02\_1000 & \_0.01\_500 & \\
			\hline
			Percentage & 2.339 & 2.640 & 2.999 & 3.492 & 5.324 & 7.232 & 8.007 & 8.601 & 9.154 & 9.345 & 9.030 & 9.848 & 10.792 & 11.034 & 11.659 & 12.125 & {1.739} & 11.642 & 11.629 & \textbf{14.657} & 10.886 & 4.074\\
			  \rowcolor{gray!30}Recall@1: m.i. & 21.184 & 20.401 & 18.957 & 17.894 & 13.280 & 9.188 & 8.044 & 7.703 & 7.422 & 7.202 & 7.342 & 6.219 & 5.978 & 6.038 & 5.256 & 5.115 & 24.453 & 11.836 & 11.675 & \textbf{4.173} & 6.620 & 15.266\\
			Recall@1: o.i. & 21.966 & 22.046 & 20.642 & 19.779 & 15.426 & 12.217 & 10.953 & 10.732 & 9.288 & 9.188 & 9.749 & 8.706 & 7.964 & 7.382 & 6.881 & \textbf{6.640} & 24.955 & 12.899 & 12.758 & 7.442 & 9.829 & 10.532\\
			  \rowcolor{gray!30}Recall@3: m.i. & 55.426 & 52.477 & 48.646 & 44.273 & 30.030 & 22.628 & 19.318 & 19.017 & 17.593 & 17.432 & 17.633 & 15.065 & 13.942 & 13.581 & 12.999 & 11.635 & 63.71 & 24.594 & 23.109 & \textbf{9.910 }& 16.489 & 37.733\\
			Recall@3: o.i. & 57.392 & 54.483 & 50.692 & 46.479 & 34.142 & 25.436 & 22.487 & 21.605 & 20.762 & 19.418 & 20.622 & 18.114 & 16.309 & 14.824 & 13.781 & \textbf{13.260} & 64.975 & 25.336 & 24.935 & 14.002 & 21.083 & 28.706\\
			  \rowcolor{gray!30}Recall@5: m.i. & 69.468 & 65.476 & 60.883 & 55.065 & 38.255 & 28.445 & 25.697 & 24.393 & 22.788 & 22.447 & 22.949 & 19.639 & 18.335 & 17.553 & 17.111 & 15.165 & 78.034 & 29.408 & 28.084 &\textbf{ 12.839} & 22.066 & 48.004\\
			Recall@5: o.i. & 70.973 & 67.161 & 63.430 & 57.633 & 41.906 & 32.016 & 28.245 & 27.081 & 25.015 & 24.333 & 25.176 & 22.066 & 21.184 & 18.776 & 17.452 & \textbf{16.830} & 78.656 & 30.652 & 30.090 & {17.212} & 26.580 & 37.813 \\
			  \rowcolor{gray!30}Recall@10: m.i. & 75.165 & 71.434 & 67.523 & 62.568 & 45.938 & 35.426 & 32.497 & 31.133 & 28.907 & 28.044 & 29.248 & 26.038 & 23.952 & 23.170 & 22.106 & 19.960 & 81.825 & 34.704 & 33.420 & \textbf{17.854} & 27.422 & 55.366\\
			Recall@10: o.i. & 76.209 & 73.240 & 69.609 & 64.574 & 50.441 & 39.699 & 35.827 & 34.363 & 31.956 & 31.234 & 32.357 & 28.425 & 27.643 & 24.774 & 23.771 & \textbf{22.487} & 82.648 & 35.807 & 35.687 & 23.049 & 32.899 & 44.433\\
			  \rowcolor{gray!30}Recall@50: m.i. & 84.072 & 82.046 & 79.418 & 75.366 & 62.969 & 52.879 & 49.789 & 47.442 & 44.353 & 44.654 & 45.537 & 42.387 & 39.799 & 37.673 & 36.429 & 35.807 & 88.104 & 48.104 & 46.941 & \textbf{30.993} & 42.828 & 70.090\\
			Recall@50: o.i. & 85.035 & 82.869 & 80.562 & 77.312 & 67.222 & 58.295 & 54.183 & 52.578 & 50.130 & 50.110 & 50.030 & 47.202 & 44.915 & 43.049 & 41.745 & 40.040 & 88.546 & 50.291 & 48.546 & \textbf{38.716} & 48.686 & 61.765\\
			  \rowcolor{gray!30}Recall@100: m.i. & 87.442 & 85.938 & 84.012 & 80.802 & 70.451 & 61.204 & 57.854 & 56.269 & 53.039 & 53.059 & 54.463 & 50.612 & 48.586 & 46.439 & 44.534 & 44.714 & 90.913 & 55.065 & 53.440 & \textbf{38.837} & 50.311 & 76.670\\
			Recall@100: o.i. & 88.185 & 86.419 & 84.855 & 82.146 & 74.162 & 66.419 & 62.708 & 61.284 & 59.198 & 58.154 & 58.054 & 56.028 & 53.220 & 51.815 & 50.441 & 48.967 & 90.973 & 56.630 & 55.406 & \textbf{46.620} & 56.329 & 69.087 \\
		\bottomrule \end{tabular} \end{adjustbox}
	\end{center}
\caption{\label{transfer_CC_arcface} Evaluation  on CC  dataset in the case of a transfer to another embedding: VQGAN and StyleGAN are optimised with MagFace and MobileFaceNet, and recognition is tested with ArcFace. We can see that generative methods' evaluation results for the CC dataset are very similar to the ones for the LFW dataset.}
\end{table*}

\begin{table*}[htbp]
	\begin{center}
	\begin{adjustbox}{max width=\textwidth}
		
		\begin{tabular}{@{}ccHHHHcHHHHHHHHHccHHcHc@{}}
  \rowcolor{gray!30}
			& VQGAN& VQGAN & VQGAN\_0.007\_128 & VQGAN & VQGAN& VQGAN & VQGAN & VQGAN & VQGAN & VQGAN & VQGAN & VQGAN & VQGAN & VQGAN & VQGAN& VQGAN & Fawkes & Fawkes & Fawkes & StyleGAN & StyleGAN & AMT-GAN\\
		 \rowcolor{gray!30}	  			&  \_0.003\_128 & \_0.005\_128 & \_0.007\_128 & \_0.01\_128 & \_0.02\_128 & & \_0.03\_256 & \_0.03\_512 & \_0.03\_1024 & \_0.03\_2048 & \_0.04\_128 & \_0.04\_256 & \_0.04\_512 & \_0.04\_1024 & \_0.04\_2048 & \_0.04\_4096 &  & \_{high4} & \_{high5} & \_0.02\_1000 & \_0.01\_500  & \\
			\hline
			Percentage & 6.513 & 6.876 & 7.309 & 7.942 & 10.225 & 12.668 & 14.042 & 14.950 & 15.656 & 16.155 & 15.221 & 16.854 & 18.369 & 19.366 & 20.065 & \textbf{20.632} & 5.200 & 8.511 & 8.305 & 15.798 & 8.290 & 7.708\\
			  \rowcolor{gray!30}Recall@1: m.i. & 20.542 & 20.381 & 19.619 & 18.415 & 11.956 & 7.763 & 5.697 & 4.814 & 3.771 & 3.811 & 4.413 & 3.029 & 2.247 & 1.565 & 1.444 & \textbf{1.224 }& 24.313 & 17.874 & 18.134 & 3.972 & 15.787 & 14.664\\
			Recall@1: o.i. & 19.719 & 19.458 & 19.960 & 18.977 & 15.707 & 12.638 & 10.712 & 9.689 & 9.448 & 8.485 & 8.967 & 7.222 & 6.640 & 6.078 & 5.737 &\textbf{ 5.155} & 24.875 & 17.472 & 18.114 & 8.004 & 16.068 & 12.919\\
			  \rowcolor{gray!30}Recall@3: m.i. & 53.440 & 51.334 & 48.887 & 43.811 & 28.606 & 17.673 & 13.761 & 11.394 & 9.729 & 9.288 & 10.692 & 7.803 & 6.239 & 4.995 & 4.213 &\textbf{ 3.852} & 60.100 & 37.332 & 37.553 & 9.850 & 39.378 & 36.871\\
			Recall@3: o.i. & 53.902 & 52.738 & 50.351 & 47.342 & 34.584 & 24.433 & 21.464 & 19.198 & 18.175 & 16.891 & 18.215 & 15.085 & 13.460 & 11.856 & 11.133 &\textbf{ 10.251} & 59.880 & 34.744 & 35.547 & 16.108 & 36.650 & 27.663\\
			  \rowcolor{gray!30}Recall@5: m.i. & 63.952 & 61.424 & 58.154 & 52.959 & 35.226 & 22.648 & 18.355 & 15.025 & 13.340 & 12.277 & 14.183 & 11.194 & 8.806 & 7.202 & 6.138 & \textbf{5.436 }& 70.812 & 44.012 & 44.393 & 13.159 & 47.884 & 44.835\\
			Recall@5: o.i. & 64.614 & 62.588 & 59.840 & 55.747 & 41.785 & 30.451 & 26.138 & 23.170 & 22.588 & 20.943 & 22.247 & 18.696 & 16.590 & 14.764 & 13.821 & \textbf{12.417} & 70.933 & 41.384 & 42.046 & 20.662 & 44.895 & 32.397\\
			  \rowcolor{gray!30}Recall@10: m.i. & 67.643 & 65.537 & 62.688 & 58.154 & 41.565 & 29.027 & 23.791 & 20.441 & 18.596 & 16.770 & 19.418 & 15.246 & 12.578 & 10.672 & 9.488 & \textbf{8.666 }& 74.022 & 49.809 & 50.512 & 18.415 & 53.781 & 50.973\\
			Recall@10: o.i. & 68.826 & 67.161 & 64.935 & 61.464 & 47.944 & 37.051 & 32.457 & 28.646 & 28.144 & 26.058 & 27.944 & 23.711 & 21.505 & 19.258 & {17.793} & \textbf{16.510} & 74.223 & 46.961 & 47.703 & 27.061 & 52.036 & 36.429\\
			  \rowcolor{gray!30}Recall@50: m.i. & 74.564 & 73.220 & 71.474 & 68.265 & 55.888 & 43.771 & 39.238 & 34.824 & 32.778 & 30.873 & 34.343 & 28.987 & 24.774 & 22.086 & 20.542 & \textbf{19.278 }& 79.539 & 61.966 & 62.407 & 33.280 & 65.296 & 64.092\\
			Recall@50: o.i. & 75.206 & 74.403 & 73.220 & 71.374 & 62.126 & 53.039 & 48.405 & 44.634 & 43.370 & 41.264 & 43.751 & 38.857 & 35.045 & 32.357 & 30.391 & \textbf{29.027} & 79.819 & 60.120 & 60.662 & 43.290 & 64.092 & 45.456\\
			  \rowcolor{gray!30}Recall@100: m.i. & 77.733 & 76.349 & 75.266 & 72.879 & 61.986 & 51.013 & 46.800 & 42.648 & 40.542 & 38.937 & 42.006 & 36.550 & 31.916 & 29.990 & 27.322 & \textbf{26.820 }& 82.006 & 67.242 & 67.703 & 41.204 & 70.231 & 69.649\\
			Recall@100: o.i. & 78.134 & 77.533 & 76.369 & 75.326 & 67.864 & 59.980 & 55.988 & 52.798 & 51.414 & 50.030 & 51.535 & 47.141 & 43.049 & 40.502 & 37.693 & \textbf{37.131} & 82.247 & 66.098 & 65.697 & 51.715 & 69.168 & 50.291\\
		\bottomrule \end{tabular} \end{adjustbox}
	\end{center}
\caption{\label{transfer_CC_facxmobile} Evaluation on CC dataset without transfer to another embedding: VQGAN and StyleGAN are optimised with MagFace and MobileFaceNet, and recognition is tested with MobileFaceNet. Lower recall and a higher percentage mean better privacy. We can see that generative methods evaluation results for the CC dataset are very similar to the ones for the LFW dataset.}
\end{table*}

\begin{table*}[htbp]
	\begin{center}
	\begin{adjustbox}{max width=\textwidth}
		
		\begin{tabular}{@{}ccHHHHcHHHHHHHHHccHHcHc@{}}
  \rowcolor{gray!30}
			& VQGAN& VQGAN & VQGAN\_0.007\_128 & VQGAN & VQGAN& VQGAN& VQGAN & VQGAN & VQGAN & VQGAN & VQGAN & VQGAN & VQGAN & VQGAN & VQGAN& VQGAN & Fawkes & Fawkes & Fawkes & StyleGAN & StyleGAN & AMT-GAN\\
			  			&  \_0.003\_128 & \_0.005\_128 & \_0.007\_128 & \_0.01\_128 & \_0.02\_128 & & \_0.03\_256 & \_0.03\_512 & \_0.03\_1024 & \_0.03\_2048 & \_0.04\_128 & \_0.04\_256 & \_0.04\_512 & \_0.04\_1024 & \_0.04\_2048 & \_0.04\_4096 &  & \_{high4} & \_{high5} & \_0.02\_1000 & \_0.01\_500 &  \\
			\hline
			Percentage & 3.025 & 3.074 & 3.218 & 3.383 & 4.209 & 4.970 & 5.351 & 5.495 & 5.740 & 5.760 & 5.826 & 6.158 & 6.472 & 6.684 & 6.866 & 6.915 & 2.652 & 6.439 & 6.067 & \textbf{7.650} & 5.452 & 4.518\\
			  \rowcolor{gray!30}Recall@1: m.i. & 20.582 & 19.900 & 19.579 & 18.957 & 15.727 & 12.758 & 12.237 & 11.735 & 11.174 & 12.056 & 10.812 & 10.712 & 9.890 & 9.408 & 9.388 & \textbf{7.783} & 24.012 & 13.019 & 13.280 & 8.345 & 14.744 & 16.790\\
			Recall@1: o.i. & 20.602 & 21.023 & 20.341 & 20.160 & 16.931 & 14.584 & 14.804 & 13.200 & 14.082 & 13.480 & 12.558 & 12.217 & 11.635 & 11.374 & 11.434 & 11.033 & 24.754 & 14.283 & 14.463 &\textbf{ 10.732} & 16.670 & 15.908\\
			  \rowcolor{gray!30}Recall@3: m.i. & 56.951 & 54.945 & 52.437 & 48.927 & 39.539 & 32.217 & 29.930 & 28.305 & 28.164 & 27.663 & 27.041 & 24.995 & 24.213 & 22.126 & 22.267 & \textbf{20.361} & 61.685 & 28.806 & 29.488 & 21.846 & 37.994 & 39.398\\
			Recall@3: o.i. & 56.971 & 55.607 & 52.558 & 49.408 & 38.897 & 33.079 & 31.595 & 30.331 & 29.609 & 28.947 & 28.064 & 26.499 & 24.754 & 24.193 & 23.450 & \textbf{22.949} & 62.347 & 30.010 & 29.809 & 24.092 & 39.478 & 34.303\\
			  \rowcolor{gray!30}Recall@5: m.i. & 69.930 & 67.743 & 64.835 & 61.003 & 49.689 & 40.943 & 38.335 & 36.710 & 35.948 & 35.065 & 35.246 & 31.976 & 30.973 & 28.686 & 28.786 & \textbf{27.021} & 75.165 & 36.048 & 36.389 & 28.646 & 48.325 & 49.328\\
			Recall@5: o.i. & 69.629 & 66.800 & 64.293 & 60.702 & 48.626 & 41.083 & 38.736 & 38.014 & 37.131 & 35.567 & 34.764 & 32.859 & 31.535 & 30.030 & 29.609 & \textbf{28.686} & 74.905 & 36.710 & 36.650 & 30.150 & 47.543 & 41.805\\
			  \rowcolor{gray!30}Recall@10: m.i. & 74.463 & 72.598 & 70.491 & 67.322 & 57.713 & 49.328 & 46.419 & 45.115 & 43.731 & 42.889 & 43.731 & 40.080 & 38.877 & 36.469 & 35.687 & \textbf{35.065} & 78.816 & 43.330 & 43.230 & 36.489 & 55.326 & 56.289\\
			Recall@10: o.i. & 74.183 & 72.337 & 70.271 & 66.961 & 56.369 & 49.589 & 47.161 & 45.617 & 45.677 & 43.390 & 43.129 & 41.163 & 39.178 & 37.613 & 37.111 & \textbf{36.189} & 78.696 & 43.129 & 43.751 & 37.232 & 54.504 & 47.141\\
			  \rowcolor{gray!30}Recall@50: m.i. & 82.608 & 81.625 & 80.261 & 78.796 & 72.718 & 66.379 & 64.072 & 63.470 & 61.825 & 61.344 & 61.003 & 58.215 & 56.650 & 55.226 & 54.323 & \textbf{53.521} & 85.496 & 58.957 & 59.599 & 54.644 & 69.188 & 71.414\\
			Recall@50: o.i. & 82.508 & 80.883 & 79.880 & 78.175 & 71.655 & 67.041 & 64.393 & 62.828 & 63.109 & 60.702 & 60.421 & 58.556 & 57.432 & 56.690 & 55.667 & \textbf{54.664} & 85.216 & 58.877 & 60.502 & 56.851 & 69.047 & 59.920\\
			  \rowcolor{gray!30}Recall@100: m.i. & 85.657 & 84.774 & 84.052 & 82.808 & 78.014 & 73.561 & 71.595 & 70.441 & 69.248 & 69.208 & 69.107 & 66.118 & 65.135 & 64.273 & 63.591 & 62.628 & 87.823 & 66.239 & 66.660 & \textbf{62.327} & 74.865 & 77.553\\
			Recall@100: o.i. & 85.537 & 84.514 & 83.771 & 82.287 & 77.833 & 73.400 & 70.913 & 70.772 & 70.451 & 69.188 & 68.345 & 66.098 & 65.637 & 65.135 & 64.333 & \textbf{63.591} & 87.803 & 66.379 & 67.182 & 64.754 & 75.266 & 65.236\\
		\bottomrule \end{tabular} \end{adjustbox}
	\end{center}
\caption{\label{transfer_CC_ResNet_152} Evaluation on CC dataset in the case of a transfer to another embedding: VQGAN and StyleGAN are optimised with MagFace + MobileFaceNet, and recognition is tested with ResNet\_152. Lower recall and a higher percentage mean better privacy. We can see that generative methods' evaluation results for the CC dataset are very similar to the ones for the LFW dataset.}
\end{table*}

\newpage

\end{document}